%% file: main.tex
\documentclass{article}
\usepackage[preprint]{myneurips_2025}

\usepackage[utf8]{inputenc}
\usepackage[T1]{fontenc}

\usepackage{amsfonts}
\usepackage{booktabs}
\usepackage{hyperref}
\usepackage{microtype}
\usepackage{nicefrac}
\usepackage{url}
\usepackage{xcolor}

\usepackage{algorithm}
\usepackage{algpseudocode}
\usepackage{bm}
\usepackage{caption}
\usepackage{enumitem}
\usepackage{mathtools}
\usepackage{pgfplots}
\usepackage{subcaption}
\usepackage{tikz}
\usepackage{wrapfig}

\algtext*{EndFor}
\algtext*{EndIf}
\algtext*{EndLoop}
\algtext*{EndWhile}

\setlist[itemize]{nosep,parsep=\parskip,leftmargin=*}

\DeclareMathOperator{\argmax}{argmax}

\pgfplotsset{compat=1.18}
\usepgfplotslibrary{colorbrewer}
\usepgfplotslibrary{fillbetween}

\usetikzlibrary{arrows}
\usetikzlibrary{arrows.meta}
\usetikzlibrary{automata}
\usetikzlibrary{calc}
\usetikzlibrary{decorations.markings}
\usetikzlibrary{positioning}
\usetikzlibrary{shapes}

\tikzset{every picture/.style={every node/.style={font=\sffamily}}}

\definecolor{cirl1}{HTML}{C44E52}
\definecolor{cirl2}{HTML}{4C72B0}
\definecolor{cpolimp1}{HTML}{D95F02}
\definecolor{cpolimp2}{HTML}{7570B3}
\definecolor{cpolimp3}{HTML}{1B9E77}
\definecolor{cpolimp4}{HTML}{E78AC3}
\definecolor{csim1}{HTML}{FC8D62}
\definecolor{csim2}{HTML}{66C2A5}
\definecolor{csim3}{HTML}{8DA0CB}

\newcommand{\justify}[1]{{#1\parfillskip=0pt\par}}
\renewcommand{\paragraph}[1]{\textbf{#1.}}

\title{Strategically Linked Decisions\\in Long-Term Planning and\\Reinforcement Learning}
\author{%
    Alihan H\"uy\"uk\\
    Harvard University\\
    \texttt{ahuyuk@seas.harvard.edu}\\
    \And
    Finale Doshi-Velez\\
    Harvard University\\
    \texttt{finale@seas.harvard.edu}\\
}

\begin{document}
\maketitle

\begin{abstract} 
    Long-term planning, as in reinforcement learning (RL), involves finding strategies: actions that collectively work toward a goal rather than individually optimizing their immediate outcomes. As part of a strategy, some actions are taken at the expense of short-term benefit to enable future actions with even greater returns. These actions are only advantageous if followed up by the actions they facilitate, consequently, they would not have been taken if those follow-ups were not available. In this paper, we quantify such dependencies between planned actions with \textit{strategic link scores}: the drop in the likelihood of one decision under the constraint that a follow-up decision is no longer available. We demonstrate the utility of strategic link scores through three practical applications: (i) explaining black-box RL agents by identifying strategically linked pairs among decisions they make, (ii) improving the worst-case performance of decision support systems by distinguishing whether recommended actions can be adopted as standalone improvements or whether they are strategically linked hence requiring a commitment to a broader strategy to be effective, and (iii) characterizing the planning processes of non-RL agents purely through interventions aimed at measuring strategic link scores---as an example, we consider a realistic traffic simulator and analyze through road closures the effective planning horizon of the emergent routing behavior of many drivers.
\end{abstract}

\section{Introduction}

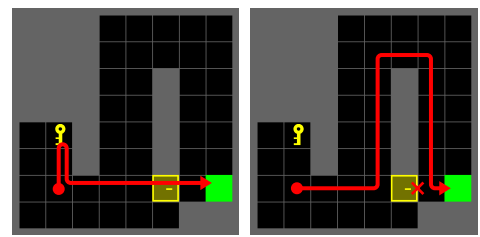
\begin{wrapfigure}[17]{r}{.5\linewidth-.5\intextsep-12pt}
    \centering
    \vspace{-\baselineskip}%
    \begin{subfigure}{.5\linewidth}
        \resizebox{\linewidth}{!}{%
            \input{figures/tikz/gridworld-policy}}%
        \caption{\bf Optimal Strategy}%
    \end{subfigure}%
    \begin{subfigure}{.5\linewidth}
        \resizebox{\linewidth}{!}{%
            \input{figures/tikz/gridworld-constraint-strategic}}%
        \caption{\bf Shortcut Blocked}%
    \end{subfigure}%
    \caption{\justify{
        \textit{An example strategy in a navigation task.} Picking up the key takes extra time early on but unlocks a major shortcut later. If the shortcut were to be blocked, going for the key would no longer be optimal, showing that picking up the key and taking the shortcut are strategically linked---the key is picked up specifically to be able to take the shortcut.}}
    \label{fig:example}
\end{wrapfigure}

A fundamental goal in reinforcement learning (RL) is to find long-term strategies---sequences of interdependent actions that work together to accomplish an objective as opposed to disjointed actions driven by their own immediate outcomes. This interdependency is a key aspect of strategic decision-making, where early actions are often taken at the expense of short-term benefit to enable future actions with even greater overall returns. When that is the case, those early actions would only be advantageous if followed up by the actions they facilitate, consequently, if an intended future action were to become unavailable with some advanced notice, a strategic decision-maker would adjust their earlier actions accordingly. In this work, we aim to formalize such strategic dependencies between decisions: 
\textit{to what extent a current action is taken to set up a future one.}

As a concrete example, consider a simple navigation task (Figure~\ref{fig:example}), where the straightforward path to the objective is long and winding. Slightly off that path, however, there is a key that unlocks a significant shortcut. In this scenario, the plan to retrieve the key---incurring a small delay---and then take the shortcut---saving more time than lost in the delay---is a strategic one. This is made evident by the fact that, if the shortcut were to somehow be blocked, an optimal agent would no longer choose to retrieve the key. With such strategic plans, we refer to ``retrieving the key'' as the \textit{set-up decision} and ``taking the shortcut'' as the \textit{pay-off decision}, which brings us to our core contribution:

\begin{itemize}[label={},leftmargin=38pt,rightmargin=38pt]
    \item We quantify the strength of the strategic link between the set-up decision and the pay-off decision as \textit{the drop in the likelihood of the set-up decision under the constraint that the pay-off decision is no longer available}, and we call this quantity the \textit{strategic link score} between the two decisions.
\end{itemize}

This drop would be large when the set-up decision is not advantageous on its own but rather made primarily to facilitate the pay-off decision. In contrast, if the optimality of the two decisions is not contingent on each other--- that is, the ``set-up'' decision is advantageous regardless of whether it is followed up by the ``pay-off'' decision---then the drop would be zero.
By providing a quantitative definition of strategic links, we enable a range of practical use cases. In this paper, we highlight three applications and demonstrate each through experiments:

\paragraph{Providing Planning-Level Explanations}
Strategic links can help explain black-box policies learned by RL algorithms. Existing explainability methods typically provide either state-level explanations---for instance, important features of a state that most influence the corresponding action \citep{greydanus2018visualizing}---or policy-level explanations---for instance, key trajectories that summarize a policy's behavior \citep{amir2018highlights}. Meanwhile, strategic links offer planning-level explanations, as they are not the result of a given policy alone but instead reflect the broader planning process and how that policy would have differed if certain decisions had been unavailable---in other words, strategic links depend on counterfactual policies as well.

\looseness-1
Given an RL agent, we identify strategic links between its decisions by searching for set-up and pay-off pairs with high strategic link scores along a trajectory of likely decisions. When a goal-driven agent's behavior is only observable through demonstrations---like human experts---we show that one can still identify strategically linked decisions by first inferring a reward function---via inverse reinforcement learning (IRL)---that captures the agent's underlying objective and serves as a model of their planning process. While reward functions learned through IRL are typically used for imitation learning---that is, to recover the demonstrated policy \citep[e.g.][]{abbeel2004apprenticeship,ziebart2008maximum}---our application highlights they are also effective at generating counterfactual policies to compute strategic link scores.

\paragraph{Improving Policies Safely via Strategy-Aware Recommendations}
Decision support systems that aim to improve policies can be made safer by determining whether recommended actions are strategically linked. Suppose a system recommends two changes to a decision-maker's policy. If the changes are strategically linked---meaning one is only effective when the other is also adopted---there is a risk that the decision-maker may unknowingly implement just one and end up worse off as a result. In such cases, the system should ideally communicate that the changes form a single strategy and must be implemented together or not at all. Conversely, if the changes are strategically independent---meaning each one is effective on its own merit regardless of whether the other is adopted or not---the decision-maker can safely choose which recommendations they want to implement. Through experiments, we show that providing strategy-aware recommendations leads to greater improvements, both on average and in worst-case scenarios.

\looseness-1
\paragraph{Characterizing Planning Behavior Through Interventions}
Strategic link scores are not limited to RL agents or reward-based planners, and they offer a means of characterizing how any agent forms long-term plans as long as it is possible to perform interventions that prevent them from taking certain actions. This is particularly useful when the agent's inner workings are inaccessible otherwise. For instance, suppose we want to understand how far ahead an agent plans. If it were an RL-based planner, we could have directly examined its discount factor. But regardless of what type of planner they are, we can still analyze strategic links between their decisions---specifically, the time distance between strongly linked decisions---to gain insight into their effective planning horizon.
We demonstrate this process in a realistic traffic simulator, where many drivers make individual routing decisions that collectively produce an emergent routing policy---one that determines how traffic flow is distributed throughout the road network. By performing interventions in the form of road closures, we characterize the implicit planning horizon of this policy and highlight how it differs from an RL-based policy.

\section{Related Work}
\label{sec:related-work}
\vspace{-6pt}

Applications of strategic links are related to \textit{explainability in RL}, which can be address at state, policy, or planning levels, and \textit{safe policy improvement}.

\paragraph{State-Level Explanations}
At this level, explanations focus on \textit{individual} decisions: why a particular state is mapped to its corresponding action. A common approach is to adapt techniques from super\-vised settings, such as extracting saliency maps \citep[e.g.][]{greydanus2018visualizing,iyer2018transparency}, collecting human annotations \citep[e.g.][]{ehsan2018rationalization}, fitting white-box meta-models \citep[e.g.][]{liu2018toward}, or generating counterfactual examples \citep[e.g.][]{olson2021counterfactual,chen2022explain,huber2023ganterfactual}. While this last group of works consider \textit{counterfactual states}, strategic links involve \textit{counterfactual policies} that could have been optimal had the environment conditions been different.

\paragraph{Policy-Level Explanations}
These explanations consider \textit{all} the decisions of a given policy collectively. Some methods summarize those decisions via key trajectories \citep{amir2018highlights} or policy graphs \citep{topin2019generation}; others contrast the policy with a baseline \citep{van2018contrastive,yao2022policy}; and many aim to learn white-box policies directly \citep[e.g.][]{khan2009minimal,shu2018hierarchical,verma2018programmatically,hein2018interpretable,silva2020optimization,sun2023accountability}. All of these methods are limited in scope to a single policy and ignore the planning process behind it. In contrast, planning-level explanations such as strategic links consider policies in context of others related through a shared planning process.

\paragraph{Planning-Level Explanations}
In RL settings, the planning processes are largely characterized by the reward function being optimized and the objective it encodes. Accordingly, a few methods explain policies by relating actions to particular dimensions of the reward function \citep{juozapaitis2019explainable,erwig2018explaining} or to the impact they have on the probability of success \citep{madumal2020explainable,yau2020did,cruz2023explainable}. In doing so, these methods provide insight not only into the policy itself, but also into why maximizing a given reward or achieving a given objective requires that particular policy. Strategic links also provide planning-level explanations, however, they relate actions to other actions rather than rewards or objective. Hence, their application is not limited to RL settings like the other methods. Strategic link scores can be computed for any type of planner, as long as their planning process can be intervened on.

\paragraph{Safe Policy Improvement}
Improving upon policies already in deployment is a common use case of RL. A key challenge in this setting is safety: ensuring the new policy does not perform worse than the original when deployed. This is usually a risk caused by the poor coverage of the available data collected by the original policy. So, prior work tends to focus on solutions at the training stage---for instance, limiting deviations from the original policy to keep new policies within coverage region \citep{laroche2019safe,wu2022supported,sharma2024decision}. Strategy-aware recommendations, on the other hand, address the post-training risk of improper or partial adoption of new policies by flagging changes that are strategically linked and should be implemented together.

\vspace{-6pt}
\section{Preliminaries}
\label{sec:notation}
\vspace{-6pt}

\paragraph{Environments}
We consider decision-making environments $\mathcal{E}=(\mathcal{S},\mathcal{A},\sigma,\tau)$, where $\mathcal{S}$ is the \textit{state} space, $\mathcal{A}$ is the \textit{action} space, $\sigma\in\Delta(\mathcal{S})$ is the \textit{initial state distribution}, and $\tau\in\Delta(\mathcal{S})^{\mathcal{S}\times \mathcal{A}}$ is the \textit{transition dynamics}. At the start of each interaction, the environment is in some initial state $s_1\sim\sigma$, and at each time step $t\in\{1,2,\ldots\}$, it transitions into a new state $s_{t+1}\sim\tau(s_t,a_t)$ depending on the current state $s_t\in\mathcal{S}$ and the action $a_t\in\mathcal{A}$ taken by a decision-maker.

\paragraph{Policies \& Planners}
Decision-makers take their actions by planning policies and then executing those policies. \textit{Policies}~$\pi\in\Delta(\mathcal{A})^\mathcal{S}$ map each state into an action distribution such that $a_t\sim\pi(s_t)$ when the policy is executed in an environment, and $\pi(a|s)$ is the probability of taking action~$a$ at state~$s$. \textit{Planners}~$\smash{\mathcal{P}^{\dagger}}$ output policies given an environment~$\mathcal{E}$, which we denote with $\smash{\pi^{\dagger}}=\smash{\mathcal{P}^{\dagger}}(\mathcal{E})$. For instance, given a \textit{reward function} $r\in\mathbb{R}^{\mathcal{S}\times\mathcal{A}}$ and a \textit{discount factor} $\gamma\in(0,1)$, the \textit{optimal planner}~$\smash{\mathcal{P}^{*(r,\gamma)}}$ aims to find the policy that maximizes the expected cumulative discounted rewards~$R$:
\begin{align}
    \pi^{*(r,\gamma)}=\mathcal{P}^{*(r,\gamma)}(\mathcal{E}) = \argmax{\!}_{\pi} ~ \mathbb{E}{}_{\pi,\mathcal{E}}[~R\doteq {\textstyle\sum_{t=1}^{\infty}}\gamma^{t-1}r(s_t,a_t)~]
\end{align}
Since our definition of strategic links involves policies planned under constraints, we denote with $\smash{\pi^{\dagger:\mathcal{C}}}=\smash{\mathcal{P}^{\dagger:\mathcal{C}}}(\mathcal{E})$ the policy planned by $\mathcal{P}^{\dagger}$ under a given constraint~$\mathcal{C}$. For instance, when states and actions are discrete, a key constraint we consider is preventing a particular action~$\tilde{a}\in\mathcal{A}$ from being taken at a particular state~$\tilde{s}\in\mathcal{S}$ such that $\pi(\tilde{a}|\tilde{s})=0$. Under that constraint, we can express the optimal planner's new policy as $\smash{\pi^{*:\{\pi(\tilde{a}|\tilde{s})=0\}}}=\smash{\mathcal{P}^{*:\{\pi(\tilde{a}|\tilde{s})=0\}}}(\mathcal{E})=\argmax{}_{\pi:\pi(\tilde{a}|\tilde{s})=0}\mathbb{E}{}_{\pi,\mathcal{E}}[R]$.

\section{Strategic Link Scores}
\label{sec:definition}

When an agent takes an action~$a\in\mathcal{A}$ at some state~$s\in\mathcal{S}$, we consider this to be a \textit{decision}---that is, decisions are state-action pairs~$(s,a)\in\mathcal{S}\times\mathcal{A}$. Then, for a given planner $\mathcal{P}^{\dagger}$, the \textit{strategic link score} between \textit{set-up decision}~$(s,a)$ and \textit{pay-off decision}~$(\tilde{s},\tilde{a})$ is defined as
\begin{align}
    \mathfrak{S}^{\dagger}_{(s,a)\to(\tilde{s},\tilde{a})} = \pi^{\dagger}(a|s) - \pi^{\dagger:\{\pi(\tilde{a}|\tilde{s})=0\}}(a|s) \label{eqn:definition}
\end{align}
which is the drop in the likelihood of the set-up, $\pi(a|s)$, when the pay-off is constrained, $\pi(\tilde{a}|\tilde{s})=0$.

\paragraph{Continuous States \& Actions}
If either the state space or the action space is continuous, this definition becomes problematic as a point constraint such as $\pi(\tilde{a}|\tilde{s})=0$ would practically have no impact on the planned policy $\smash{\pi^{\dagger}}$. In such cases, we allow a region of pay-off decisions, $\smash{\tilde{\mathcal{S}}}\times\smash{\tilde{\mathcal{A}}}\subseteq\mathcal{S}\times\mathcal{A}$, and work with the constraint $\pi(\tilde{a}|\tilde{s})=0, \forall\tilde{s}\in\smash{\tilde{\mathcal{S}}}, \forall\tilde{a}\in\smash{\tilde{\mathcal{A}}}$.

\looseness-1
The choice of a meaningful region will depend on the application. For instance, the traffic scenario we consider in Section~\ref{sec:application3} involves determining the (continuous) traffic flow $a\in\mathcal{A}=[0,1]$ through junctions $s\in\mathcal{S}=\{1,\ldots,10\}$. If we are interested in whether the flow through the first junction~$s=1$ is strategically dependent on sufficient flow at the last junction~$s=10$, we can consider the pay-off region: $\smash{\tilde{\mathcal{S}}}=\{10\}$ and $\smash{\tilde{\mathcal{A}}}=(a^*,1]$ for some threshold~$a^*$. This encodes the constraint that the flow $a\sim\pi(s=10)$ cannot be larger than the threshold---that is $a\leq a^*$ almost surely.

\begin{wrapfigure}[12]{r}{.5\linewidth-.5\intextsep-48pt}
    \centering
    \vspace{-\baselineskip-3pt}%
    \begin{tikzpicture}[
        ->,
        >=stealth,
        shorten >=1pt,
        node distance=2cm,
        semithick,
        state/.append style={minimum size=0,inner sep=2pt}]
    
        \node[state] (S1) {$S_1$};
        \node[state,right of=S1] (S2) {$S_2$};
        \path ($(S1.west)+(-0.5cm,0)$) edge node {} (S1);
        \path (S1) edge[loop above] node[left,xshift=-0.125cm,yshift=0.125cm,align=right] {$A_1$\\[-3pt]\tiny($r_{\alpha}=1$)\\[-3pt]\tiny($r_{\beta}=1$)} (S1);
        \path (S2) edge[loop above] node[left,xshift=-0.125cm,yshift=0.125cm,align=right] {$A_1$\\[-3pt]\tiny($r_{\alpha}=1$)\\[-3pt]\tiny($r_{\beta}=1$)} (S2);
        \path (S1) edge node[below,align=center] {$A_2$\\[-3pt]\color{cirl1}\tiny($\bm{r_{\alpha}=0.5}$)\\[-3pt]\color{cirl2}\tiny($\bm{r_{\beta}=1.5}$)} (S2);
        \path (S2) edge[loop right] node[below,xshift=0.25cm,align=center] {$A_2$\\[-3pt]\color{cirl1}\tiny($\bm{r_{\alpha}=2.5}$)\\[-3pt]\color{cirl2}\tiny($\bm{r_{\beta}=1.5}$)} (S2);
        
    \end{tikzpicture}%
    \vspace{-3pt}%
    \caption{\textit{Dynamics of the toy example}. For initial state $S_1$, the action sequence $(A_2,A_2)$ is optimal under both reward functions $\smash{r_{\alpha}}$ and $\smash{r_{\beta}}$. For $\smash{r_{\alpha}}$, the two actions are strategically linked as $A_2|S_1$ is not optimal unless $A_2|S_2$. For $\smash{r_{\beta}}$, there is no strategic link as $A_2|S_1$ is optimal regardless of the action taken at state~$S_2$.}%
    \label{fig:toyexample}
\end{wrapfigure}

\paragraph{A Toy Example}
Consider a simple environment with two states $\mathcal{S}=\{S_1,S_2\}$ and two actions $\mathcal{A}=\{A_1,A_2\}$. The transitions are deterministic as shown in Figure~\ref{fig:toyexample} and the initial state is always $S_1$. We are given two different reward functions, $r_{\alpha}$ and $r_{\beta}$, also shown in Figure~\ref{fig:toyexample}. Suppose two optimal planners, $\mathcal{P}^{*(r_\alpha)}$ and $\mathcal{P}^{*(r_{\beta})}$, aim to each maximize one of these reward functions, for a fixed time horizon of two steps and no discounting.

In either case, the optimal policy is exactly the same: take action $A_2$ at $S_1$ and then $A_2$ at $S_2$, which can be expressed as
\begin{align}
    \pi^{*(r_{\alpha})}(S_1)=\pi^{*(r_{\beta})}(S_1)=A_2 \\
    \pi^{*(r_{\alpha})}(S_2)=\pi^{*(r_{\beta})}(S_2)=A_2
\end{align}

Despite having the same policies, for planner $\mathcal{P}^{*(r_{\alpha})}$, these two decisions are strategically linked. Under $r_{\alpha}$, the immediate reward of $A_1$ is larger than $A_2$ at $S_1$: $r_{\alpha}(S_1,A_1)=1>r_{\alpha}(S_1,A_2)=0.5$. However, $\mathcal{P}^{*(r_{\alpha})}$ still takes action $A_2$ to be able to transition into $S_2$, where an even larger pay-off is available, namely $r_{\alpha}(S_2,A_2)=2.5$ over $r_{\alpha}(S_1,A_1)=1$. If taking action $A_2$ at $S_2$ had not been an option, taking action $A_2$ at $S_1$ would no longer have been optimal, therefore
\begin{align}
    \mathfrak{S}^{*(r_{\alpha})}_{(S_1,A_2)\to(S_2,A_2)} \:=\: \underbrace{\pi^{*(r_{\alpha})}(A_2|S_1)}_{~~=\,1} \:-\: \underbrace{\pi^{*(r_{\alpha}):\{\pi(A_2|S_2)=0\}}(A_2|S_1)}_{~~=\,0} \:=\: 1
\end{align}
Meanwhile, for planner $\mathcal{P}^{*(r_{\beta})}$, there is no strategic link. Under $r_{\beta}$, transitioning into $S_2$ is not important as both $A_1$ and $A_2$ have the same reward in either state: $r_{\beta}(S_1,A_1)=r_{\beta}(S_2,A_1)=1$ and $r_{\beta}(S_1,A_2)=r_{\beta}(S_2,A_2)=1.5$. Taking action $A_2$ at $S_1$ is optimal regardless of what action is taken at the next state $S_2$, therefore
\begin{align}
    \mathfrak{S}^{*(r_{\beta})}_{(S_1,A_2)\to(S_2,A_2)} \:=\: \underbrace{\pi^{*(r_{\beta})}(A_2|S_1)}_{~~=\,1} \:-\: \underbrace{\pi^{*(r_{\beta}):\{\pi(A_2|S_2)=0\}}(A_2|S_1)}_{~~=\,1} \:=\: 0
\end{align}
\looseness-1
This toy example underlines why strategic link scores lead to planning-level explanations: While policies $\pi^{*(r_{\alpha})}$ and $\pi^{*(r_{\beta})}$ are the same, they exhibit different strategic links depending on the planner, $\mathcal{P}^{*(r_{\alpha})}$ or $\mathcal{P}^{*(r_{\beta})}$. It also underlines why such strategic analysis is important for safe policy improvement. Suppose the current policy is suboptimal and always takes action~$A_1$ with total value $r(S_1,A_1)+r(S_1,A_1)=2$ under either reward function. Then, suppose we recommend the optimal policy instead: taking action~$A_2$ first at state~$S_1$ and then at state~$S_2$. Under $r_{\alpha}$, when the recommendations are strategically linked, implementing the first one without implementing the second would result in worse performance: $r_{\alpha}(S_1,A_2)+r_{\alpha}(S_2,A_1)=1.5 < 2$. Under $r_{\beta}$, when there is no strategic link, the same implementation would still improve performance: $r_{\beta}(S_1,A_2)+r_{\beta}(S_2,A_1)=2.5 > 2$.

\vspace{-6pt}
\section{Practical Applications}
\vspace{-6pt}

\looseness-1
Having defined the strategic link score $\mathfrak{S}$, we now illustrate three of its applications introduced earlier: providing planning-level explanations (Section~\ref{sec:application1}), improving policies safely via strategy-aware recommendations (Section~\ref{sec:application2}), and characterizing planning behavior through interventions (Section~\ref{sec:application3}).  We present empirical results for each application, with detailed setup given in the appendix.

\vspace{-6pt}
\subsection{Providing Planning-Level Explanations}
\label{sec:application1}
\vspace{-3pt}

\begin{wrapfigure}[11]{r}{.5\linewidth-.5\intextsep-48pt}
    \vspace{-\baselineskip-20pt}%
    \begin{minipage}{\linewidth}
        \vspace{-\baselineskip}%
        \begin{algorithm}[H]
            \small
            \captionsetup{font=small}
            \algrenewcommand\alglinenumber[1]{\scriptsize #1:}
            \caption{\\Planning-Level Explanations}
            \label{alg:explanation}
            \begin{algorithmic}[1]
                \State \textbf{Input:} Planner $\smash{\mathcal{P}^{\dagger}}$
                \State $\smash{s_1\sim \sigma_1}$
                \For{$\smash{t\in\{1,\ldots,T\}}$}
                    \State $\smash{a_t\gets \argmax_{a\in\mathcal{A}}\pi^{\dagger}(a|s_t)}$
                    \State $\smash{s_{t+1}\sim \tau(s_t,a_t)}$
                \EndFor
                \For{$\smash{t\in\{1,\ldots,T\}}$}
                    \For{$\smash{t'\in\{t,\ldots,T\}}$}
                        \State $\bar{\mathfrak{S}}^{\dagger}_{tt'}\gets\mathfrak{S}^{\dagger}_{(s_t,a_t)\to(s_{t'},a_{t'})}$
                        \vspace{-2pt}
                    \EndFor
                \EndFor
                \State \textbf{Output:} Scores $\smash{\bar{\mathfrak{S}}}$
            \end{algorithmic}%
        \end{algorithm}
    \end{minipage}
\end{wrapfigure}

\paragraph{Setup}
We consider grid-based maze environments, called \texttt{GridWorld}, as in our introductory example in Figure~\ref{fig:example}. In these environments, the agent is tasked with navigating to a target cell, and they incur a constant negative reward each time step they have not reached the target. As actions, the agent can move either up, down, left, or right, and by first walking over keys, they can later pass through previously locked doors. In this application, we aim to explain policies learned by soft value iteration \citep{haarnoja2017reinforcement} for solving various maze layouts. We make sure none of the layouts become unsolvable if a particular decision is constrained.

\paragraph{Approach 1: Links from Known Planner}
Given a planner $\mathcal{P}^{\dagger}$, we explain the policy~$\pi^{\dagger}$ by first selecting a trajectory representative of its behavior, and then computing the strategic link score between each pair of decisions along that trajectory. We consider trajectories, rather than random decisions, as they are more likely to exhibit strategic links. While any trajectory of interest can be analyzed in the same way, for this demonstration, we use the most likely trajectory, obtained by taking the highest-probability action at each step. When computing scores, we treat the earlier decision in each pair as the set-up and the latter as the pay-off. The complete procedure is given in Algorithm~\ref{alg:explanation}.

\begin{figure}
    \centering
    \begin{subfigure}{.25\linewidth-9pt}
        \resizebox{\linewidth}{!}{%
            \input{figures/tikz/gridworld-policy}}%
        \vspace{6pt}%
        \caption{\bf Optimal Trajectory \\~}%
        \label{fig:app1-simple-panel1}
    \end{subfigure}%
    \begin{subfigure}{.25\linewidth-9pt}
        \resizebox{\linewidth}{!}{%
            \input{figures/tikz/gridworld-constraint-strategic}}%
        \vspace{6pt}%
        \caption{\bf Constraint with\\High Strategic Links}%
        \label{fig:app1-simple-panel2}
    \end{subfigure}%
    \begin{subfigure}{.25\linewidth-9pt}
        \resizebox{\linewidth}{!}{%
            \input{figures/tikz/gridworld-constraint-greedy}}%
        \vspace{6pt}%
        \caption{\bf Constraint with\\Low Strategic Links}%
        \label{fig:app1-simple-panel3}
    \end{subfigure}%
    \hspace{3pt}%
    \begin{subfigure}{.25\linewidth+24pt}
        \resizebox{\linewidth}{!}{%
            \input{figures/tikz/gridworld-scores}}%
        \vspace{-3.9pt}
        \caption{\bf Strategic Link Scores\\along the Optimal Trajectory}%
        \label{fig:app1-simple-panel4}
    \end{subfigure}%
    \vspace{-3pt}%
    \caption{\textit{Strategic link scores for a simple maze layout.} Creating a shortcut by picking up the key to unlocking the door is strategic (a), since blocking the door results in the key not being picked up (b), while constraining an unrelated action does not lead to the same outcome (c). By looking at the strategic link scores between all the decisions along the optimal trajectory (d), the link between the key (``K'') and the door (``D'') can be seen clearly.}%
    \label{fig:app1-simple}%
    \vspace{-.5\baselineskip}
    \vspace{-6pt}%
\end{figure}
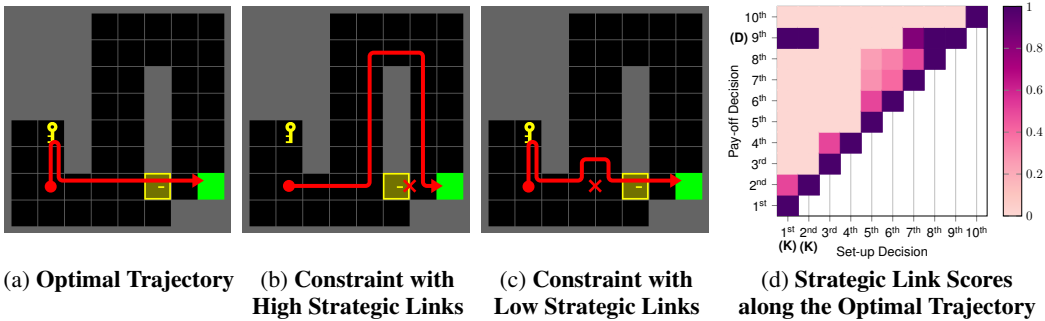

\looseness-1
\paragraph{Results for A Simple Layout}
Figure~\ref{fig:app1-simple} shows our approach in action for the maze layout from our introduction. In Figure~\ref{fig:app1-simple-panel1}, we plot the most likely trajectory under the soft optimal policy: The target can be reached via a long and winding path, but a quicker strategy is to pick up a key that is slightly out of the way to unlock a door, creating a significant shortcut. We know the key and the door are strategically linked because if we block the door, picking up the key is no longer optimal, see Figure~\ref{fig:app1-simple-panel2}. Preventing another decision that is not related to the key does not lead to the same outcome, see Figure~\ref{fig:app1-simple-panel3}. In Figure~\ref{fig:app1-simple-panel4}, we plot the strategic link scores between all potential set-up and pay-off decisions along the optimal trajectory, which makes it clear that moving towards the key (1\textsuperscript{st} and 2\textsuperscript{nd} decisions, labeled ``K'') are indeed a set-up for going through the door (9\textsuperscript{th} decision, labeled ``D'').

\begin{figure}
    \centering
    \begin{minipage}{.5\linewidth-1pt}
        \centering
        \begin{subfigure}{.5\linewidth-18pt-1pt}
            \resizebox{\linewidth}{!}{%
                \input{figures/tikz/gridworld-unlinked-policy}} \\
            \resizebox{\linewidth}{!}{%
                \input{figures/tikz/gridworld-unlinked-constraint}}%
            \vspace{2pt}%
            \caption{\bf Layout with\\Independent Keys}%
            \label{fig:app1-twokeys-unlinked-layout}
        \end{subfigure}%
        \hspace{2pt}%
        \begin{subfigure}{.5\linewidth+18pt-1pt}
            \resizebox{\linewidth}{!}{%
                \input{figures/tikz/gridworld-unlinked-scores}}%
            \vspace{-5pt}%
            \caption{\bf Link Scores for\\Independent Keys}%
            \label{fig:app1-twokeys-unlinked-scores}
        \end{subfigure}%
    \end{minipage}%
    \hspace{2pt}%
    \begin{minipage}{.5\linewidth-1pt}
        \centering
        \begin{subfigure}{.5\linewidth-18pt-1pt}
            \resizebox{\linewidth}{!}{%
                \input{figures/tikz/gridworld-linked-policy}} \\
            \resizebox{\linewidth}{!}{%
                \input{figures/tikz/gridworld-linked-constraint}}%
            \vspace{2pt}%
            \caption{\bf Layout with Correlated Keys}%
            \label{fig:app1-twokeys-linked-layout}
        \end{subfigure}%
        \hspace{2pt}%
        \begin{subfigure}{.5\linewidth+18pt-1pt}
            \resizebox{\linewidth}{!}{%
                \input{figures/tikz/gridworld-linked-scores}}%
            \vspace{-5pt}%
            \caption{\bf Link Scores for\\Correlated Keys}%
            \label{fig:app1-twokeys-linked-scores}
        \end{subfigure}%
    \end{minipage}%
    \vspace{-3pt}%
    \caption{\textit{Strategic link scores for layout with independent vs.\ correlated keys.} When the keys unlock separate shortcuts (a), they is no strategically link between them (b). If one key is skipped, collecting the other still remains optimal. When the keys jointly unlock a single shortcut (c), the decisions to pick up each key are strategically linked (d). If one key is skipped, collecting the other becomes pointless.}%
    \label{fig:app1-twokeys}
    \vspace{-4pt}%
\end{figure}
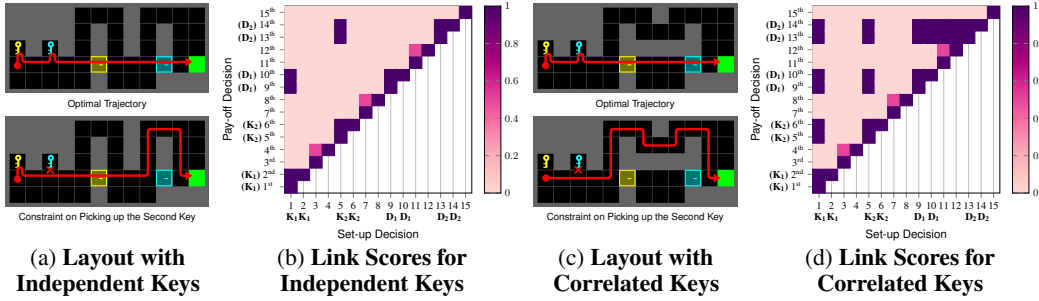

\paragraph{Results for Linked \& Unlinked Keys}
In Figure~\ref{fig:app1-twokeys}, we consider two other layouts, each with an additional key and its corresponding door. In the first layout (Figure~\ref{fig:app1-twokeys-unlinked-layout}), the two keys unlock separate shortcuts, so whether one of them is retrieved or not has no effect on the optimality of retrieving the other. In Figure~\ref{fig:app1-twokeys-unlinked-scores}, we see that each key (``K\textsubscript{1}'' and ``K\textsubscript{2}'') is strategically linked to its corresponding door (``D\textsubscript{1}'' and ``D\textsubscript{2}''), but not to the other key-door pair. In the second layout (Figure~\ref{fig:app1-twokeys-unlinked-layout}), there is just a single shortcut and unlocking it requires both keys to be collected. Now, the keys are strategically linked: If one of them is not retrieved, there is no point in retrieving the other (seen in Figure~\ref{fig:app1-twokeys-linked-scores}).

These two layouts also highlight the importance of strategy-aware recommendations. If an agent is not taking advantage of any shortcuts by collecting keys, we might recommend them to pick up both keys. In the independent case, it is safe for the agent to only follow through on just collecting one of the recommended keys. In the strategically linked case, however, if the agent were to collect only one key without the other, they would unnecessarily waste time as the shortcut would remain inaccessible.

\begin{figure}
    \centering
    \begin{subfigure}{.33\linewidth-2pt}
        \centering
        \scalebox{.44}{%
            \input{figures/tikz/irl-gridworld}}%
        \vspace{-3pt}%
        \caption{\texttt{GridWorld}}%
    \end{subfigure}%
    \hspace{3pt}%
    \begin{subfigure}{.33\linewidth-2pt}
        \centering
        \scalebox{.44}{%
            \input{figures/tikz/irl-shortcuts}}%
        \vspace{-3pt}%
        \caption{\texttt{Shortcuts}}%
    \end{subfigure}%
    \hspace{3pt}%
    \begin{subfigure}{.33\linewidth-2pt}
        \centering
        \scalebox{.44}{%
            \input{figures/tikz/irl-arthighway}}%
            \vspace{-3pt}%
        \caption{\texttt{ArterialHighway}}%
    \end{subfigure}%
    \vspace{-3pt}%
    \caption{\textit{Strategic link scores inferred from demonstrations} become more accurate with increasing variation, following a similar trend to reward inference (until policies are almost uniformly random and rewards become unidentifiable, strategic link scores remain accurate as recognizing policies to be uniformly random is sufficient).}%
    \label{fig:app1-inverse}%
    \vspace{-\baselineskip-4pt}%
\end{figure}
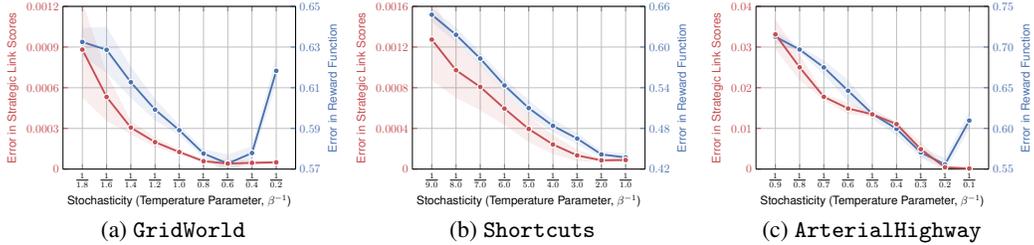

\looseness-1
\paragraph{Approach 2: Links from Demonstrations}
When a goal-driven planner is not available as an explicit algorithm but only observable through demonstrations---that is, trajectories generated by its unconstrained policy---we can still identify strategic links by first inferring a reward function that captures the planner's objective. We achieve this via maximum entorpy IRL \citep{ziebart2008maximum}. Because the coverage of demonstrations is critical for IRL, we experiment with different levels of stochasticity in actions by varying the temperature parameter in soft value iteration. We evaluate inferred rewards using the EPIC distance, \citep{gleave2020quantifying}, which accounts for equivalent reward expressions up to shaping, and measure the accuracy of strategic link scores computed based on those rewards using mean square error (MSE). In addition to \texttt{Gridworld}, we also consider two other environments from later applications: \texttt{Shortcuts}, abstract environments with similar key-shortcut dynamics (Section~\ref{sec:application2}), and \texttt{ArterialHighway}, a realistic traffic simulation (Secion~\ref{sec:application3}).
We run each experiment five times with different seeds to obtain 1-sigma error bars.

In Figure~\ref{fig:app1-inverse}, we see that strategic link scores inferred from demonstrations improve in accuracy as the variation in those demonstrations increases. This is similar to the trend with inferred reward functions with one notable difference: When policies are so stochastic that they essentially pick actions uniformly at random, reward functions become unidentifiable, hence the sudden spikes in the corresponding error. However, strategic link scores are unaffected by this issue as recognizing policies to be uniformly random---without finding the exact reward structure---happens to be sufficient to correctly conclude that no decision is strategic. These results show that demonstrations alone can carry enough information to model not just the policy of a decision-maker but also their planning process in the form of a reward function, and thereby identify their strategies.

\vspace{-6pt}
\subsection{Improving Policies Safely via Strategy-Aware Recommendations}
\label{sec:application2}
\vspace{-3pt}

\paragraph{Environment}
For this application, we introduce \texttt{Shortcuts}, a procedurally-generated collection of abstract environments that are designed to have dynamics similar to key-shortcut relationships in from our previous application with more complex strategic links between different shortcuts.

\begin{wrapfigure}[33]{r}{.5\linewidth-.5\intextsep}
    \centering
    \resizebox{\linewidth}{!}{%
        \input{figures/tikz/shortcuts}}%
    \caption{\justify{%
        \textit{An example environment that has 5~nodes, 3~shortcuts, and 4~preparation actions.} After taking the required preparation actions, the agent can jump via shortcuts to move more efficiently towards the target.}}%
    \label{fig:app2-environment}

    \vspace{12pt}
    \resizebox{\linewidth}{!}{%
        \input{figures/tikz/policy-improvement}}%
    \caption{\textit{Performance following various recommendation methods.} Pick-and-Choose is not safe, potentially leading to worse performance than to begin with. All-or-Nothing is safe but not effective unless a large number of recommendations are implemented. Strategic-Aware is both safe and effective for smaller number of implemented recommendations.}%
    \label{fig:app2-result}
\end{wrapfigure}

\looseness-1
Figure~\ref{fig:app2-environment} shows an example \texttt{Shortcuts} environment. The agent starts at some initial node and needs to move one node at a time until they reach the $N$-th node. Each move costs one unit of time but reaching the target returns a reward of $N$. What makes the environment interesting is shortcuts that jump over multiple nodes at once. Each shortcut becomes available only after performing certain preparation actions beforehand (similar to keys in \texttt{GridWorld}). These preparation actions do not cause the agent to move but they still cost a fixed amount $C<1/2$. They become advantageous through shortcuts they enable: Jumping forward via a shortcut that spans $n$ nodes with $k$ required preparations costs $n-k\cdot 2C$ rather than $n$ individual moves normally needed. We generate 100 environments with 10 nodes, 5 shortcuts, and 5 preparation actions, setting $C=0.1$. Which nodes the shortcuts jump over and which preparations they require are randomized, while preparation actions are always placed at the initial node.

\paragraph{Objective}
Our goal is to recommend actions to a suboptimal agent to improve their performance. We consider an agent that knows of shortcuts and makes use of them when they are available, however, their behavior suboptimal because they are not aware of the required preparations---that is, they never take a preparation action unless one is recommended by us. Hence, a recommendation is essentially a set of specific preparation actions. For instance, in Figure~\ref{fig:app2-environment}, the two shorter jumps happen to be more efficient, so the optimal recommendations are $\{\texttt{prep}_1,\texttt{prep}_2,\texttt{prep}_4\}$.

\begin{wrapfigure}[11]{r}{.5\linewidth-.5\intextsep-48pt}
    \vspace{-\baselineskip}%
    \begin{minipage}{\linewidth}
        \vspace{-\baselineskip-1pt}%
        \begin{algorithm}[H]
            \small
            \captionsetup{font=small}
            \algrenewcommand\alglinenumber[1]{\scriptsize #1:}
            \caption{\\Strategy-Aware Recommendations}
            \label{alg:recommendation}
            \begin{algorithmic}[1]
                \State \textbf{Input:} Planner $\smash{\mathcal{P}^{\dagger}}\!$, its recommended decisions $\smash{\mathcal{D}=\{(s,a)_i\}_{i=1}^N}$
                \For{$\smash{i\in\{1,\ldots,N\}}$}
                    \State $\smash{\mathcal{D}_i\gets\{(s,a)_i\}}$
                    \For{$\smash{j\in\{1,\ldots,N\}}$}
                        \If{$\mathfrak{S}^{\dagger}_{(s,a)_i\to(s,a)_j}\gg 0$}
                            \vspace{3pt}
                            \State $\smash{\mathcal{D}_i\gets\mathcal{D}_i\cup\{(s,a)_j\}}$
                        \EndIf
                    \EndFor
                \EndFor
                \State \textbf{Output:} $\smash{\{\mathcal{D}_1,\ldots,\mathcal{D}_N\}}$
            \end{algorithmic}%
        \end{algorithm}
        \vspace{--1pt}%
    \end{minipage}
\end{wrapfigure}

\paragraph{Methods}
Given a set of recommendations, we are interested how to best present them to the agent, and we consider three approaches.
(i) \textbf{Pick-and-Choose}:~One can present them directly without any explanation, allowing the agent to pick and choose which recommendations they want to implement. During evaluation, we consider all possible combinations and report average as well as worst-case performances. This approach has a risk: If two recommendations are strategically linked, the agent might unknowingly implement only one and end up worse off as a consequence. 
(ii) \textbf{All-or-Nothing}:~One can inform the agent that the recommendations might be related, and hence should either be implemented all together or not at all. This can be overly cautious: If the agent is not willing to make large changes in their behavior, they might miss out on individual improvements that are not reliant on other recommendations.
We propose (iii) \textbf{Strategy-Aware}:~Each recommendation is considered as a potential set-up decision, and it is grouped together with every other recommendation that is strategically linked to it as a pay-off decision (see Algorithm~\ref{alg:recommendation}). As an example, consider again the recommendations for the environment in Figure~\ref{fig:app2-environment}. \textit{Strategy-Aware} would group them as $\{~\{\texttt{prep}_1,\texttt{prep}_2\},\allowbreak~\{\texttt{prep}_4\}~\}$, while \textit{Pick-and-Choose} is equivalent to the grouping $\{~\{\texttt{prep}_1\},~\{\texttt{prep}_2\},~\{\texttt{prep}_4\}~\}$, and \textit{All-or-Nothing} to the grouping $\{~\{\texttt{prep}_1,\allowbreak\texttt{prep}_2,\allowbreak\texttt{prep}_4\}~\}$.

\paragraph{Results}
For each strategy, we consider all subsets of recommendations the agent can implement without breaking groups, record the average as well as the worst-case performance, and take their mean across the 100 random environments we have procedurally generated. Figure~\ref{fig:app2-result} shows the results broken down with respect to the number of individual recommendations implemented. We see that, indeed, the worst-case performance of \textit{Pick and Choose} tends to be worse than that of the original policy. \textit{All-or-Nothing} is safer, never resulting in worse performance, as it does not allow for partial adoption of recommendations. However, it leads to significant improvements only if a large number of recommendations are implemented. Meanwhile, \textit{Strategy-Aware}, which takes advantage of our concept of strategically-linked actions, is safe and also effective even when the agent is only willing to make small changes to their policy.

\subsection{Characterizing Planning Behavior Through Interventions}
\label{sec:application3}

Strategic link scores offer a way to characterize how agents plan their behavior purely through in\-ter\-ven\-tion\-al data, without needing to know anything about their decision-making process. We demonstrate this by examining driving behavior in a realistic traffic simulator, \texttt{UXsim} \citep{seo2025joss}. Our goal is to understand how far ahead drivers plan when making routing decisions. Although the simulator has explicit rules for an individual driver's behavior, how these rules---executed by many drivers---collectively shape the traffic flow through a road network is difficult to track formally. Instead, we rely only on interventions and use them to measure strategic link scores.

\begin{wrapfigure}[13]{r}{.5\linewidth-.5\intextsep}
    \centering
    \vspace{-\baselineskip-3pt}%
    \resizebox{\linewidth}{!}{%
        \input{figures/tikz/arthighway}}%
    \vspace{-3pt}
    \caption{\justify{\textit{The traffic scenario.} `Entry' to `exit', drivers need to decide whether to stay on an arterial road or divert to the highway at junctions `J1' to `J10'. Despite having a lower speed limit, the arterial route is shorter hence overall quicker. But, if one of its segments were to be closed off, it is better to divert to the highway the soonest to make use of its higher speed limit for longer.}}
    \label{fig:app3-environment}
\end{wrapfigure}
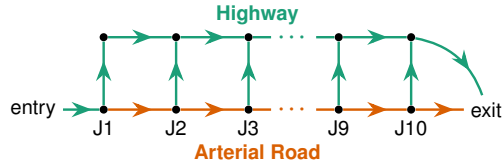

\paragraph{Road Network}
We consider the road network in Figure~\ref{fig:app3-environment}, which we call \texttt{ArterialHighway}. Each driver enters the network through the same road and aims to reach the same exit, either using a highway or an arterial road. Along the arterial road, there are junctions, ``J1'' to ``J10'', where drivers need to decide whether to stay on the arterial road or switch to the highway via on-ramps. The highway has a higher speed limit but it is overall a longer route as the on-ramps add to the distance. Under no congestion, the shorter arterial route still takes less time despite the lower speed limit, hence staying on the arterial road at every junction is the optimal policy. However, if any segment of the arterial road were to be closed off, then drivers would have no choice but to eventually divert to the highway, and to take advantage of the higher speed limit for longer, diverting as early as possible would become optimal. This pressure to act early makes this road network particularly suitable for measuring planning horizon.

\paragraph{Intervention}
Taking advantage of the road structure, we can measure the planning horizon of drivers through a simple intervention: closing off the arterial road at the last junction and then observing how early drivers tend to change their route. What we measure is the strategic link score between routing decisions at earlier junctions, as set-ups, and the decision at the last junction as the pay-off. In formal terms, let $\pi^{\dagger}:\mathcal{S}=\{\text{J1},\ldots,\text{J10}\}\to\mathcal{A}=[0,1]$ be the collective policy that emerges out of individual drivers' decisions such that $\pi^{\dagger}(\text{JX})$ is the frequency with which drivers take the arterial road at junction JX. Then, we are interested in the following strategic link score, modified slightly to fit the continuous action setting: $\smash{\mathfrak{S}_{\text{JX}\to\text{J10}}^{\dagger}}=\pi^{\dagger}(\text{JX})-\smash{\pi^{\dagger:\{\pi(\text{J10})=0\}}}(\text{JX})$.

\begin{wrapfigure}[17]{r}{.5\linewidth-.5\intextsep}
    \centering
    \vspace{-\baselineskip-0.5pt}
    \resizebox{\linewidth}{!}{%
        \input{figures/tikz/cars-sim-vehicles}}%
    \vspace{-5pt}%
    \caption{\textit{Vehicle counts.} The rate at which vehicles pass through each junction and still stay on the arterial road changes significantly after the intervention at J10.}%
    \label{fig:app3-simulation-panel1}
\end{wrapfigure}
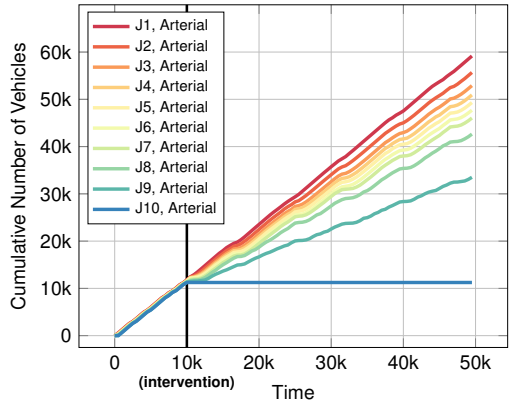

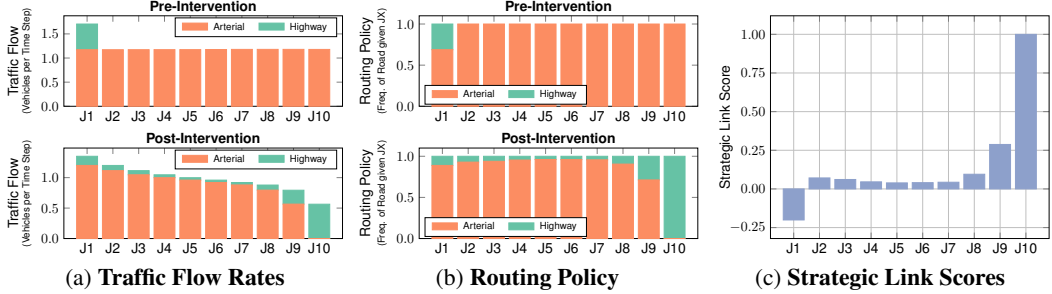
\begin{figure}
    \centering
    \begin{subfigure}{.333\linewidth-2pt}
        \raggedright
        \scalebox{.527}{%
            \input{figures/tikz/cars-sim-flow}}%
        \vspace{-3pt}%
        \caption{\bf Traffic Flow Rates}%
        \label{fig:app3-simulation-panel2}
    \end{subfigure}%
    \hspace{3pt}%
    \begin{subfigure}{.333\linewidth-2pt}
        \centering
        \scalebox{.527}{%
            \input{figures/tikz/cars-sim-policy}}%
        \vspace{-3pt}%
        \caption{\bf Routing Policy}%
        \label{fig:app3-simulation-panel3}
    \end{subfigure}%
    \hspace{3pt}%
    \begin{subfigure}{.333\linewidth-2pt}
        \raggedleft
        \scalebox{.527}{%
            \input{figures/tikz/cars-sim-strategy}}%
        \vspace{-3pt}%
        \caption{\bf Strategic Link Scores}%
        \label{fig:app3-simulation-panel4}
    \end{subfigure}%
    \vspace{-3pt}%
    \caption{\textit{Strategic link scores for the simulated drivers.} Using the count data in Figure~\ref{fig:app3-simulation-panel1}, we extract traffic flow rates (a), and normalizing those rates, we obtain the emergent routing policy of drivers, pre- and post-intervention (b). Strategic link scores are the difference between the two policies (c). When it comes to the decision of saying on the arterial road, the strongest strategic link to J10---besides J10 itself---is at J9. In other words, following the closure of the arterial road past J10, the drivers tend to divert to the highway mostly at J9.}%
    \label{fig:app3-simulation}%
    \vspace{-12pt}
\end{figure}

\paragraph{Simulation}
We run a simulation for 50k time steps, where the intervention to close off the arterial road at J10 is performed at time step 10k. Figure~\ref{fig:app3-simulation-panel1} shows our raw data: the number of vehicles over time that pass through each junction and stay on the arterial road; we see a clear change in behavior after the intervention. By taking the average slope of these count-over-time plots, we obtain traffic flow rates---number of vehicles per time step---at each junction, before and after the intervention, either staying on the arterial road or switching to the highway (Figure~\ref{fig:app3-simulation-panel2}). By normalizing the flow rates at each junction so that they add up to one, we obtain the pre- and post-intervention routing policies (Figure~\ref{fig:app3-simulation-panel3}). Taking the difference between the two policies gives us the strategic link scores we are after (Figure~\ref{fig:app3-simulation-panel4}).

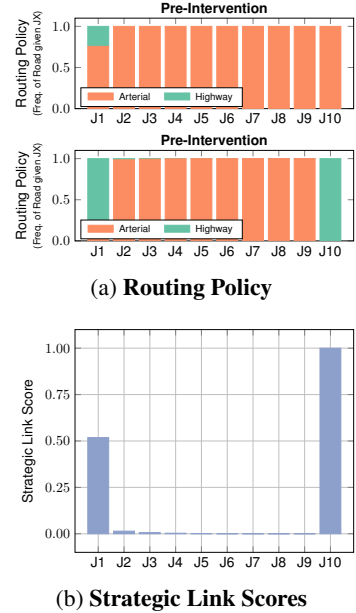
\begin{wrapfigure}[26]{r}{.5\linewidth-.5\intextsep-60pt}
    \centering
    \vspace{-3pt}%
    \begin{subfigure}{\linewidth}
        \raggedleft
        \scalebox{.527}{%
            \input{figures/tikz/cars-rl-policy}}%
        \caption{\bf Routing Policy}%
    \end{subfigure}%

    \vspace{9pt}
    \begin{subfigure}{\linewidth}
        \raggedleft
        \scalebox{.527}{%
            \input{figures/tikz/cars-rl-strategy}}%
        \caption{\bf Strategic Link Scores}%
    \end{subfigure}%
    \caption{\textit{Strategic link scores for the optimal routing behavior.} Unlike the behavior of the simulated drivers, the optimal response to the road closure past J10 would have been at J1.}%
    \label{fig:app3-reinforcement}
\end{wrapfigure}

\paragraph{Results}
First, we see that the link score for J10 is one. This is by definition of link scores; every decision is strategically linked to itself with a score one. Since the intervention is performed at J10, every driver reaching J10 has no choice but to continue to the highway. Besides J10, we see that the most significant link score is at J9, corresponding to a 25\% percentage-point drop in the rate of drivers that still decide to stay on the arterial road and continue to J10. This suggest that the collective behavior of drivers respond rather myopically to changes in traffic conditions. Compare their behavior to the optimal routing policy shown in Figure~\ref{fig:app3-reinforcement}, which we compute via RL and would only be possible if the drivers were to all cooperate. Besides J10, the only strategic link is to J1---the earliest diversion point---which requires a planning horizon of at least ten junctions.

\looseness-1
While the emergent behavior of drivers is limited in horizon when diverting from the arterial road, one long-horizon effect still stands out: the significant \textit{negative} link score at J1. This reflects drivers who were \textit{avoiding} the arterial road even before the intervention, and whose \textit{avoidance} was strategically linked to the road connection following J10. That connection is what made majority of the drivers prefer the arterial road, leading to congestion that slowed down the traffic enough for the longer highway route to become competitive. That is why some drivers decided to avoid the arterial road altogether. Once the intervention caused drivers to leave the arterial road, reducing congestion, this avoidance strategy became less frequent---hence the negative strategic link score.

\vspace{-6pt}
\section{Conclusion}
\vspace{-6pt}

We introduced \textit{strategic link scores}, which formalize a key aspect of long-term planning: actions being taken to enable future actions. By identifying strategic links, we were able to provide planning-level explanations of RL agents, and even directly improve decision-making performance by guiding safe implementation of action recommendations. Beyond reward-based settings, we also used strategic link scores as a tool for characterizing planning processes that are given as explicit algorithms.

\paragraph{Limitations}
While the definition in \eqref{eqn:definition} works well in discrete settings, it may need to be interpreted differently in continuous settings depending on environment structure and objective---for instance, our final application provided one interpretation tailored to the traffic scenario we analyzed. Moreover, not all strategic links are captured by \eqref{eqn:definition}; some strategies may involve higher-order dependencies: If A is taken to enable either B or C, constraints put on B or C only would not reveal the full relationship.

\paragraph{Societal Impact}
Strategic links are a general concept in planning, not tied to a specific application. Our strategy-aware approach to decision support is one way they can positively impact society.

\clearpage
\bibliographystyle{myIEEEtranSN}
\bibliography{references}

\clearpage
\appendix

\section{Experimental Details}

This appendix outlines the implementation details of our experiments. All experiments were run on a personal computer with 64GB of RAM and no GPUs. Each individual experiment took less than a couple hours to complete.

\subsection{Details for Providing Planning-Level Explanations}

\paragraph{Approach 1: Links from Known Planner}
In \texttt{GridWorld}, the state space consists of the agent's position, represented as row and column indices~$(i,j)$, and a binary flag~${f_k}$ for each key~$k$, indicating whether the corresponding key has been retrieved. The action space consists of the four cardinal directions: up, down, left, and right. Transitions are deterministic. At each time step when the agent is not in the target cell, they receives a reward of $-1$. When collecting the most likely trajectory in Algorithm~\ref{alg:explanation}, we let the agent to interact with the environment until they reach the target, which determines the value of $T$.

As the planner, we consider soft value iteration \citep{haarnoja2017reinforcement}, setting the discount factor as $\gamma=0.99$, the number of iterations as $250$, and the inverse temperature parameter as $\beta=100$. When computing strategic link scores, we constrain the decision~$(s,a)$ by setting the reward value at that decision to negative infinity: $r(s,a)\gets -\infty$. In \texttt{GridWorld}, we ignore flags~$\{f_k\}$ and only consider the agent's position~$(i,j)$ to be part of the state. That is, we consider moving a specific direction in a specific cell to be the same decision, regardless of which keys have been picked up so far. Accordingly, when constraining the decision~$(i,j,\{f_k\},a)$, we set $r(i,j,\{f'_k\},a)\gets\infty$ for all $f'_k\in\{0,1\}$, independent of the values~$\{f_k\}$.

\paragraph{Approach 2: Links from Demonstrations}
For these experiments, we compute strategic link scores using reward functions inferred from demonstrations via IRL, rather than using the true reward functions. The specific scores we evaluate depends on the environment: For \texttt{GridWorld}, we consider the same score matrices as in ``Links from Known Planner''; for \texttt{Shorcuts}, we consider the same scores that needed to be computed for the application in Section~\ref{sec:application2}; and similarly for \texttt{ArterialHighway}, we consider the same scores computed in Section~\ref{sec:application3}.

Although compute the same scores as in our applications, we use slightly smaller versions of each environments to reduce the computational demand of performing IRL: For \texttt{GridWorld}, we consider only the simple layout shown in Figure~\ref{fig:app1-simple}; for \texttt{Shortcuts}, we generate $10$ environments with $5$ nodes, $3$ shortcuts, and $3$ preparation actions instead of $100$ environments with $10$ nodes, $5$ shortcuts, and $5$ preparation actions; for \texttt{ArterialHighway}, we consider a road network with $5$ junctions instead of $10$. In \texttt{ArterialHighway}, using soft value iteration requires us to quantize the continuous action space. We normally use $100$ quantization points, but for these experiments, we reduce it to $10$.

As demonstrations, we sample $10,\!000$ trajectories following optimal policies computed via soft value iteration. In \texttt{GridWorld}, we fix the time horizon of each trajectory to $T=W\times H$, where $W$ is the grid width and $H$ is the grid height; in \texttt{Shortcuts}, we fix $T=N\times L$, where $N$ is the number of nodes and $L$ is the number of preparation actions; and in \texttt{ArterialHighway}, we fix $T=J+1$, where $J$ is the number of junctions. For each environment, the number of soft value iteration steps is set accordingly. We vary the inverse temperature parameter to obtain demonstrations with different levels of stochasticity---see Figure~\ref{fig:app1-inverse} for the specific temperature values considered.

As the IRL algorithm, we use maximum entropy IRL \citep{ziebart2008maximum}. The RL algorithm used in the inner loop of maximum entropy IRLis again soft value iteration with the same hyperparameters (number of iterations and inverse temperature) as those used for generating demonstrations. Since all our environments have deterministic transitions, we consider the setting where the true transition dynamics are known. Meanwhile, reward functions are represented as $|\mathcal{S}|$-by-$|\mathcal{A}|$ matrices; initialized to all zeros; and updated via gradient steps over $10,\!000$ iterations using the Adam optimizer with hyperparameters $\beta_1=0.9$, $\beta_2=0.999$, and $\epsilon=10^{-8}$. The learning rate is set to $10^{-4}$ for \texttt{GridWorld} and \texttt{Shortcuts}, and to $5\cdot 10^{-3}$ for \texttt{ArterialHighway}.

\vspace{-6pt}
\subsection{Details for Improving Policies Safely via Strategy-Aware Recommendations}
\vspace{-3pt}

In \texttt{Shortcuts}, the state space consists of the node that the agent is currently at, along with binary flags corresponding to each preparation action, indicating whether that action has been taken or not. The action space consists of moving forward (\texttt{move}), jumping via the $i$-th shortcut ($\{\texttt{jump}_i\}$), and or taking the $j$-th preparation action ($\texttt{prep}_j$). Letting $n$ be the current node and $f_j$ be the flag corresponding to action $\texttt{prep}_j$, the transition dynamics are given by the following rules:
\begin{alignat}{7}
    n&\gets n+1 \quad&&\text{if}\quad a=\texttt{move} &&\wedge \mathcal{V}_{\texttt{move}} &&\doteq \{n\neq N\} \\
    n&\gets n_{\texttt{jump}_i}^{\text{(to)}} \quad&&\text{if}\quad a=\texttt{jump}_j &&\wedge \mathcal{V}_{\texttt{jump}_i} &&\doteq \{n=n_{\texttt{jump}_i}^{\text{(from)}}\} \wedge \{f_j=1, \forall j\in \mathcal{J}_{\texttt{jump}_i}\} \\
    f_j&\gets1 \quad&&\text{if}\quad a=\texttt{prep}_j &&\wedge \mathcal{V}_{\texttt{prep}_j} &&\doteq \{n=n_{\texttt{prep}_j}\}
\end{alignat}
where $\mathcal{V}_a$ denotes the validity condition for action~$a$, $n_{\texttt{jump}_i}^{\text{(to)}}$ is the destination of shortcut~$i$, $n_{\texttt{jump}_i}^{\text{(from)}}$ is the origin of shortcut~$i$, $\mathcal{J}_{\texttt{jump}_i}$ is the set of preparation actions required to use shortcut~$i$, and $n_{\texttt{prep}_j}$ is the node where action~$\texttt{prep}_j$ is located at. Meanwhile, the reward function is given by
\begin{align}
    r(n,\{f_j\},\texttt{move}) &= \begin{cases}-1&\text{if}\quad n+1\neq N\\-1+N &\text{if}\quad n+1=N\end{cases} \\
    r(n,\{f_j\},\texttt{jump}_i) &= \begin{cases}-1&\text{if}\quad\neg\mathcal{V}_{\texttt{jump}_i}\\-(n_{\texttt{jump}_i}^{\text{(to)}}-n_{\texttt{jump}_i}^{\text{(from)}})+|\mathcal{J}_{\texttt{jump}_i}|\cdot2C&\text{if}\quad\mathcal{V}_{\texttt{jump}_i}\wedge n_{\texttt{jump}_i}^{\text{(to)}}\neq N\\-(n_{\texttt{jump}_i}^{\text{(to)}}-n_{\texttt{jump}_i}^{\text{(from)}})+|\mathcal{J}_{\texttt{jump}_i}|\cdot2C+N &\text{if}\quad\mathcal{V}_{\texttt{jump}_i}\wedge n_{\texttt{jump}_i}^{\text{(to)}}=N\end{cases} \\
    r(n,\{f_j\},\texttt{prep}_j) &= \begin{cases}-1&\text{if}\quad\neg\mathcal{V}_{\texttt{prep}_j}\\-C&\text{if}\quad\mathcal{V}_{\texttt{prep}_j}\end{cases}
\end{align}

We generate these environments randomly according to following procedure, letting $I$ be the number of shortcuts and $J$ be the number of preparation actions:
\begin{algorithm}[H]
    \caption{\texttt{Shortcuts} Environment Generation}
    \begin{algorithmic}[1]
        \For{$i\in\{1,\ldots,I\}$}
            \State $n'\sim\text{Uniform}(\{1,\ldots,N\})$; $n''\sim\text{Uniform}(\{1,\ldots,N\}\setminus\{n'\})$
            \State $n_{\texttt{jump}_i}^{\text{(from)}}\gets \min\{n',n''\}$; $n_{\texttt{jump}_i}^{\text{(to)}}\gets \max\{n',n''\}$
            \State $J'\sim\text{Uniform}(\{1,\ldots,J\})$; $\mathcal{J}_{\texttt{jump}_i}\gets \{\}$
            \For{$\_\in\{1,\ldots,J'\}$}
                \State $j'\sim\text{Uniform}(\{1,\ldots,J\}\setminus\mathcal{J}_{\texttt{jump}_i})$; $\mathcal{J}_{\texttt{jump}_i}\gets\mathcal{J}_{\texttt{jump}_i}\cup\{j'\}$
            \EndFor
        \EndFor
        \For{$j\in\{1,\ldots,J\}$}
            \State $n_{\texttt{prep}_j}\gets 1$
        \EndFor
    \end{algorithmic}
\end{algorithm}
\vspace{-.5\baselineskip}
When using soft value iteration to plan policies, we set the discount factor as $\gamma=0.99$, the number of iterations as $30$ (twice the total number of nodes and preparation actions), and the inverse temperature as $\beta=100$. When checking whether a strategic link score is significantly greater than zero, as in Line~5 of Algorithm~\ref{alg:recommendation}, we set the significance threshold to halfway between $0$ and $1/5$ (one over the number of preparation action). This is because: When all preparation actions are recommended together, a soft-optimal policy assigns them equal probabilities at the initial state, since the order in which they are taken does not matter. Therefore, the strategic link score between any two preparation actions becomes at most $1/5$, as the probability of a recommendation cannot drop more than its original value after a constraint.

\subsection{Details for Characterizing Planning Behavior Through Interventions}

\paragraph{Simulation}
We refer to \citet{seo2025joss} for the details regarding the mechanisms underlying the traffic simulator. In our experiments, the simulation parameters are set as follows: $\texttt{deltan}=5$, $\texttt{reaction\_time}=1$, $\texttt{duo\_update\_time}=500$, $\texttt{duo\_update\_weight}=0.5$, and $\texttt{duo\_noise}=0.01$. Each road segment along the highway, as well as the on-ramps leading to it, has two lanes, a length of $1000$, and a free flow speed of $20$. Road segments along the arterial have the same number of lanes and the same length, but a lower free flow speed of $20\sqrt{J/(J+1)}$. We choose this value so that the arterial road is reasonably quicker (takes less overall time): If the speed on the arterial road were to be $20\cdot J/(J+1)$, both the arterial and highway routes would have taken the same amount of time as their length ratio is $J/(J+1)$. If it were to be $20$ instead, so that the two routes have the same speed, shorter length would have always meant a quicker route, requiring no strategic routing. The actual free flow speed is the geometric mean of these extreme scenarios. Finally, the incoming traffic flow into the road network through the ``entry'' is set to be $2.0$.

\paragraph{Reinforcement Learning}
In addition to simulated agents, we also use RL to find the optimal routing policy that minimizes the average travel time throughout the whole road network. Each road segment has a travel time that depends on the traffic flow it carries, which we denote as $T_A(f)$ and $T_H(f)$ given flow~$f$ for the arterial and the highway (including on-ramps) respectively. Since the incoming flow must be equal to the outgoing flow at each junction, the initial entry flow~$f_0$ determines the flow throughout the network, conditioned on a routing policy $\pi$.
Letting $f_{A,J_i}$, $f_{R,J_i}$, and $f_{H,J_i}$ denote the traffic flow on the arterial segment following junction~$J_i$, the on-ramp leaving that junction, and the highway segment that follows, we can write
\begin{alignat}{5}
    f_{A,J_1}&=f_0\cdot\pi(J_1) &\hspace{12pt} f_{R,J_1}&=f_0-f_{A,J_1} &\hspace{12pt} f_{H,J_1} &= f_{R,J_1} \\
    f_{A,J_i}&=f_{A,J_{i-1}}\cdot\pi(J_i) &\hspace{12pt} f_{R,J_i}&=f_{A,J_{i-1}}-f_{A,J_i} &\hspace{12pt} f_{H,J_i}&=f_{H,J_{i-1}}+f_{R,J_i} \\[-2pt] &&&&&={\textstyle\sum_{i'=1}^i}f_{R,J_{i'}} =f_0-f_{A,J_i}
\end{alignat}

Notice that the flow rates past each junction, $f_{A,J_i}$, $f_{R,J_i}$, $f_{H,J_i}$, can be determined solely from the flow rate coming into that junction, $f_{A,J_{i-1}}$. Leveraging this structure, we find the policy $\pi(J_i)$ that would result in the shortest average travel time efficiently via the following three steps:
First, we include in our state space the incoming flow~$f$ to junction~$J_i$, for which we can write the following deterministic transition function:
\begin{align}
    \tau(J_i,f,a) = (J_{i+1},f'=fa)
\end{align}
Second, we use value iteration to compute the optimal flow-dependent policy $\pi^*(J_i,f)$, which captures optimal routing under any flow condition, not just for our specific entry flow rate~$f_0$.
Third, we roll this policy out, starting from the entry flow rate~$f_0$, to obtain a flow-free policy $\pi^*(J_i)$, which would remain optimal as long as the entry flow rate does not change:
\begin{alignat}{3}
    \pi^*(J_1)&\gets\pi^*(J_1,f_0) &\hspace{24pt} f_{A,J_1}&\gets f_0\cdot \pi^*(J_1) \\
    \pi^*(J_i)&\gets\pi^*(J_i,f_{A,J_{i-1}}) &\hspace{24pt} f_{A,J_i}&\gets f_{A,J_{i-1}}\cdot \pi^*(J_i)
\end{alignat}
We use the following reward function
\begin{align}
    r(J_i,f,a) &= \underbrace{fa\cdot T_A(fa)}_{\text{arterial}} + \underbrace{f(1-a)\cdot T_H(f(1-a))}_{\text{on-ramp}} + \underbrace{(f_0-fa)\cdot T_H(f_0-fa)}_{\text{highway}}
\end{align}
which ensures that cumulative rewards correspond to the average travel time across all road segments, weighted by the actual traffic flow carried by each segment. When using value iteration, we quantize the continuous actions space $\mathcal{A}=[0,1]$ using $100$ equally-spaced quantization points.

\end{document}

%% file: figures/tikz/gridworld-policy.tex
\begin{tikzpicture}
    \node at (0,0) {\includegraphics[width=170pt]{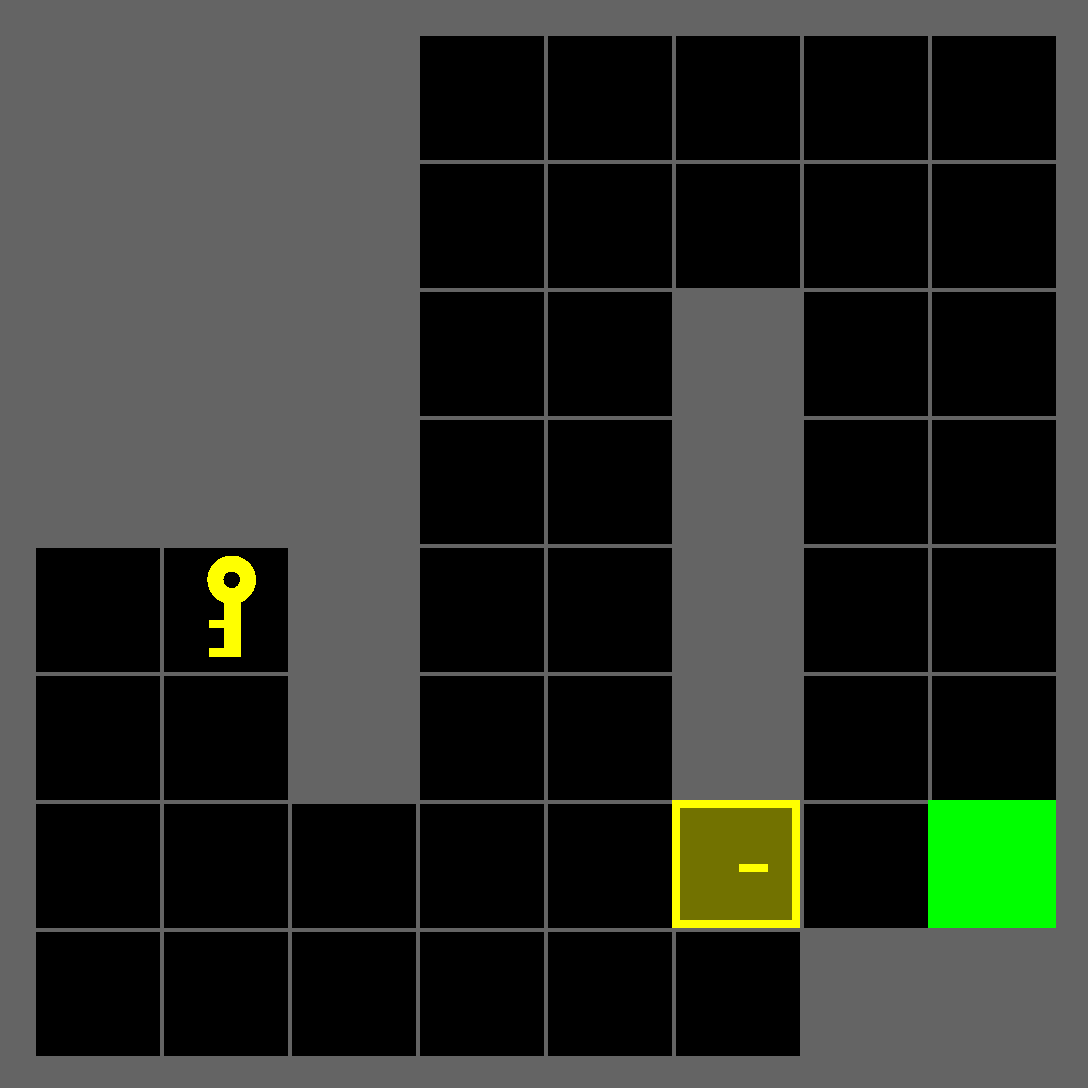}};

    \coordinate (ini) at (-50pt, -50pt);
    \coordinate (key) at (-50pt, -10pt);
    \coordinate (end) at (70pt, -50pt);

    \draw[
        {Circle[scale=0.75]}-{Triangle[scale=0.75]},
        red,
        line width=3pt,
        rounded corners=3pt]
        
        ($(ini) + (0, -5pt)$)
        -- ($(key) + (0, -7pt)$)
        -- ($(key) + (6pt, -7pt)$)
        -- ($(ini) + (6pt, 4pt)$)
        -- ($(end) + (-5pt, 4pt)$);
\end{tikzpicture}

%% file: figures/tikz/gridworld-constraint-strategic.tex
\begin{tikzpicture}
    \node at (0,0) {\includegraphics[width=170pt]{figures/raw/gridworld-wsimple}};

    \coordinate (ini) at (-50pt, -50pt);
    \coordinate (key) at (-50pt, -10pt);
    \coordinate (end) at (70pt, -50pt);

    \draw[
        {Circle[scale=0.75]}-{Triangle[scale=0.75]},
        red,
        line width=3pt,
        rounded corners=3pt]
        
        ($(ini) + (-5pt, 0)$)
        
        -- ($(10pt, -50pt) + (0, 0)$)
        -- ($(10pt, 50pt) + (0, 0)$)
        -- ($(50pt, 50pt) + (0, 0)$)
        -- ($(50pt, -50pt) + (0, 0)$)
    
        -- ($(end) + (-5pt,0)$);

    \node[
        cross out,
        draw,
        red,
        line width=2pt] at (40pt, -50pt) {};
\end{tikzpicture}

%% file: figures/tikz/gridworld-constraint-greedy.tex
\begin{tikzpicture}
    
    \node at (0,0) {\includegraphics[width=170pt]{figures/raw/gridworld-wsimple}};

    \coordinate (ini) at (-50pt, -50pt);
    \coordinate (key) at (-50pt, -10pt);
    \coordinate (end) at (70pt, -50pt);
    
    \draw[
        {Circle[scale=0.75]}-{Triangle[scale=0.75]},
        red,
        line width=3pt,
        rounded corners=3pt]
        
        ($(ini) + (0, -5pt)$)
        -- ($(key) + (0, -7pt)$)
        -- ($(key) + (6pt, -7pt)$)
        -- ($(ini) + (6pt, 4pt)$)
        
        -- ($(-10pt, -50pt) + (0, 4pt)$)
        -- ($(-10pt, -30pt) + (0, 0)$)
        -- ($(10pt, -30pt) + (0, 0)$)
        -- ($(10pt, -50pt) + (0, 4pt)$)
    
        -- ($(end) + (-5pt,4pt)$);

    \node[
        cross out,
        draw,
        red,
        line width=2pt] at (0, -50pt) {};
\end{tikzpicture}

%% file: figures/tikz/gridworld-scores.tex
\begin{tikzpicture}
    \begin{axis}[
        colormap/RdPu,
        colorbar,
        colorbar style={ymin=0},
        point meta min=-0.2,
        point meta max=1,
        axis equal image,
        xlabel={Set-up Decision},
        ylabel={Pay-off Decision},
        xlabel style={yshift=12pt},
        ylabel style={yshift=-15pt},
        xmin=-0.5, xmax=9.5,
        ymin=-0.5, ymax=9.5,
        xtick={0,1,2,3,4,5,6,7,8,9},
        xticklabels={
            \shortstack{1\textsuperscript{st}\\\textbf{(K)}},
            \shortstack{2\textsuperscript{nd}\\\textbf{(K)}},
            3\textsuperscript{rd},
            4\textsuperscript{th},
            5\textsuperscript{th},
            6\textsuperscript{th},
            7\textsuperscript{th},
            8\textsuperscript{th},
            9\textsuperscript{th},
            10\textsuperscript{th}},
        ytick={0,1,2,3,4,5,6,7,8,9},
        yticklabels={
            1\textsuperscript{st},
            2\textsuperscript{nd},
            3\textsuperscript{rd},
            4\textsuperscript{th},
            5\textsuperscript{th},
            6\textsuperscript{th},
            7\textsuperscript{th},
            8\textsuperscript{th},
            \textbf{(D)} 9\textsuperscript{th},
            10\textsuperscript{th}},
        xmajorgrids=true,
        xtick pos=bottom,
        ytick pos=left,
        tick align=outside]
        
    \addplot[
        matrix plot*,
        point meta=explicit]
        file {figures/raw/gridworld-wsimple-scores.dat};
    \end{axis}
\end{tikzpicture}

%% file: figures/tikz/gridworld-unlinked-policy.tex
\begin{tikzpicture}
    \node at (0,0) {\includegraphics[width=250pt]{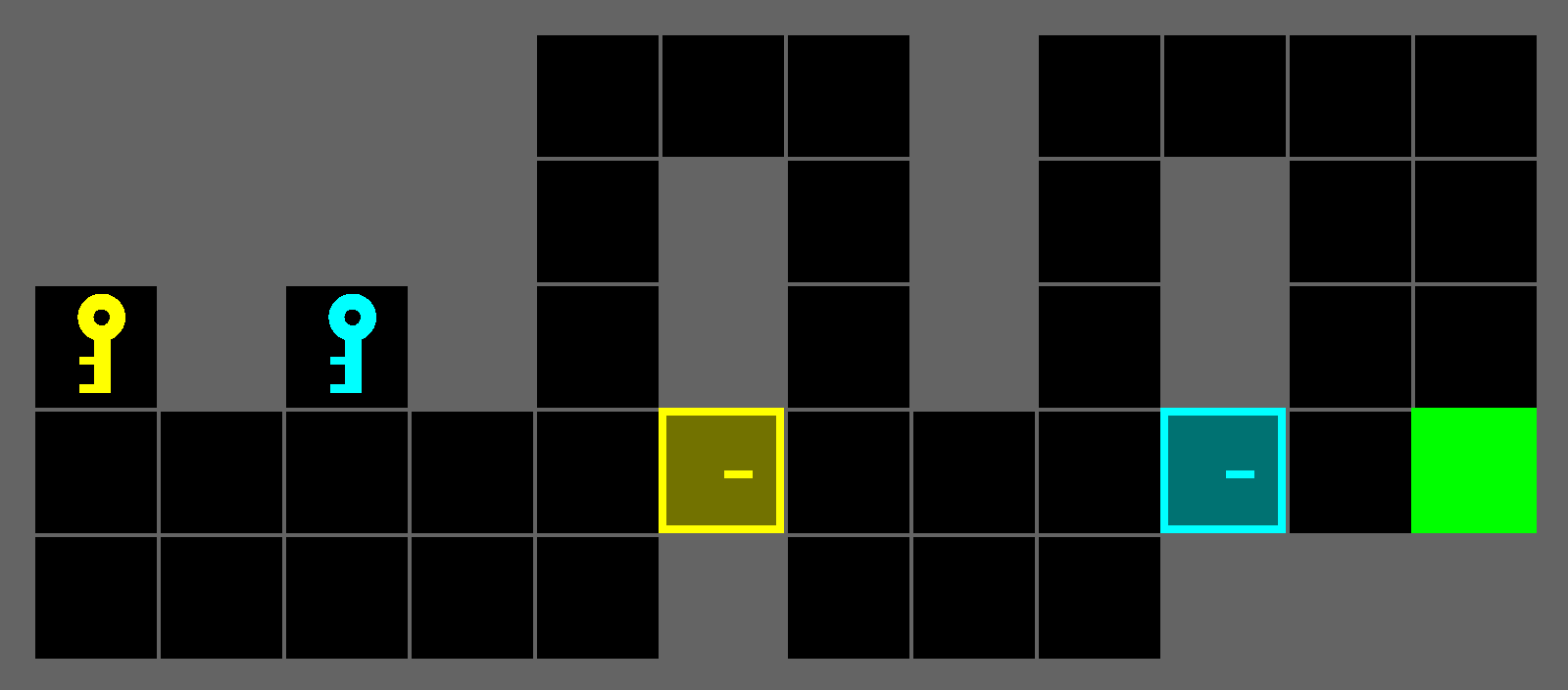}};

    \coordinate (ini) at (-110pt, -20pt);
    \coordinate (key) at (-110pt, 0pt);
    \coordinate (mni) at (-70pt, -20pt);
    \coordinate (mey) at (-70pt, 0pt);
    \coordinate (end) at (110pt, -20pt);

    \draw[
        {Circle[scale=0.75]}-{Triangle[scale=0.75]},
        red,
        line width=3pt,
        rounded corners=3pt]
        
        ($(ini) + (0, -6pt)$)
        -- ($(key) + (0, -7pt)$)
        -- ($(key) + (6pt, -7pt)$)
        -- ($(ini) + (6pt, 3pt)$)
        -- ($(mni) + (0, 3pt)$)
        -- ($(mey) + (0, -7pt)$)
        -- ($(mey) + (6pt, -7pt)$)
        -- ($(mni) + (6pt, 3pt)$)
        -- ($(end) + (-5pt, 3pt)$);

    \node[align=center] at (0, -69pt) {\large Optimal Trajectory};
\end{tikzpicture}

%% file: figures/tikz/gridworld-unlinked-constraint.tex
\begin{tikzpicture}
    \node at (0,0) {\includegraphics[width=250pt]{figures/raw/gridworld-unlinked}};

    \coordinate (ini) at (-110pt, -20pt);
    \coordinate (key) at (-110pt, 0pt);
    \coordinate (mni) at (-70pt, -20pt);
    \coordinate (mey) at (-70pt, 0pt);
    \coordinate (end) at (110pt, -20pt);

    \draw[
        {Circle[scale=0.75]}-{Triangle[scale=0.75]},
        red,
        line width=3pt,
        rounded corners=3pt]
        
        ($(ini) + (0, -6pt)$)
        -- ($(key) + (0, -7pt)$)
        -- ($(key) + (6pt, -7pt)$)
        -- ($(ini) + (6pt, 3pt)$)

            -- ($(-30pt, -20pt) + (0, 3pt)$)
            -- ($(50pt, -20pt) + (0, 3pt)$)
            -- ($(50pt, 40pt) + (0, 0)$)
            -- ($(90pt, 40pt) + (0, 0)$)
            -- ($(90pt, -20pt) + (0, 3pt)$)
            
            -- ($(end) + (-5pt,3pt)$);

    \node[
        cross out,
        draw,
        red,
        line width=2pt] at ($(mey) + (0pt, -10pt)$) {};

    \node[align=center] at (0, -69pt) {\large Constraint on Picking up the Second Key};
\end{tikzpicture}

%% file: figures/tikz/gridworld-unlinked-scores.tex
\begin{tikzpicture}
    \begin{axis}[
        colormap/RdPu,
        colorbar,
        colorbar style={ymin=0},
        point meta min=-0.2,
        point meta max=1,
        axis equal image,
        xlabel={Set-up Decision},
        ylabel={Pay-off Decision},
        ylabel style={yshift=-3pt},
        xmin=-0.5, xmax=14.5,
        ymin=-0.5, ymax=14.5,
        xtick={0,1,2,3,4,5,6,7,8,9,10,11,12,13,14},
        xticklabels={
            \shortstack{1\\\textbf{~K\textsubscript{1}}},
            \shortstack{2\\\textbf{~K\textsubscript{1}}},
            3,
            4,
            \shortstack{5\\\textbf{~K\textsubscript{2}}},
            \shortstack{6\\\textbf{~K\textsubscript{2}}},
            7,
            8,
            \shortstack{9\\\textbf{~D\textsubscript{1}}},
            \shortstack{10\\\textbf{~~D\textsubscript{1}}},
            11,
            12,
            \shortstack{13\\\textbf{~~D\textsubscript{2}}},
            \shortstack{14\\\textbf{~~D\textsubscript{2}}},
            15},
        ytick={0,1,2,3,4,5,6,7,8,9,10,11,12,13,14},
        yticklabels={
            \textbf{(K\textsubscript{1})} 1\textsuperscript{st}\hspace*{2pt},
            \textbf{(K\textsubscript{1})} 2\textsuperscript{nd},
            3\textsuperscript{rd},
            4\textsuperscript{th},
            \textbf{(K\textsubscript{2})} 5\textsuperscript{th},
            \textbf{(K\textsubscript{2})} 6\textsuperscript{th},
            7\textsuperscript{th},
            8\textsuperscript{th},
            \textbf{(D\textsubscript{1})} ~~9\textsuperscript{th},
            \textbf{(D\textsubscript{1})} 10\textsuperscript{th},
            11\textsuperscript{th},
            12\textsuperscript{th},
            \textbf{(D\textsubscript{2})} 13\textsuperscript{th},
            \textbf{(D\textsubscript{2})} 14\textsuperscript{th},
            15\textsuperscript{th}},
        tick label style={font=\footnotesize},
        xmajorgrids=true,
        xtick pos=bottom,
        ytick pos=left,
        tick align=outside]
        
    \addplot[
        matrix plot*,
        point meta=explicit]
        file {figures/raw/gridworld-unlinked-scores.dat};
    \end{axis}
\end{tikzpicture}

%% file: figures/tikz/gridworld-linked-policy.tex
\begin{tikzpicture}
    \node at (0,0) {\includegraphics[width=250pt]{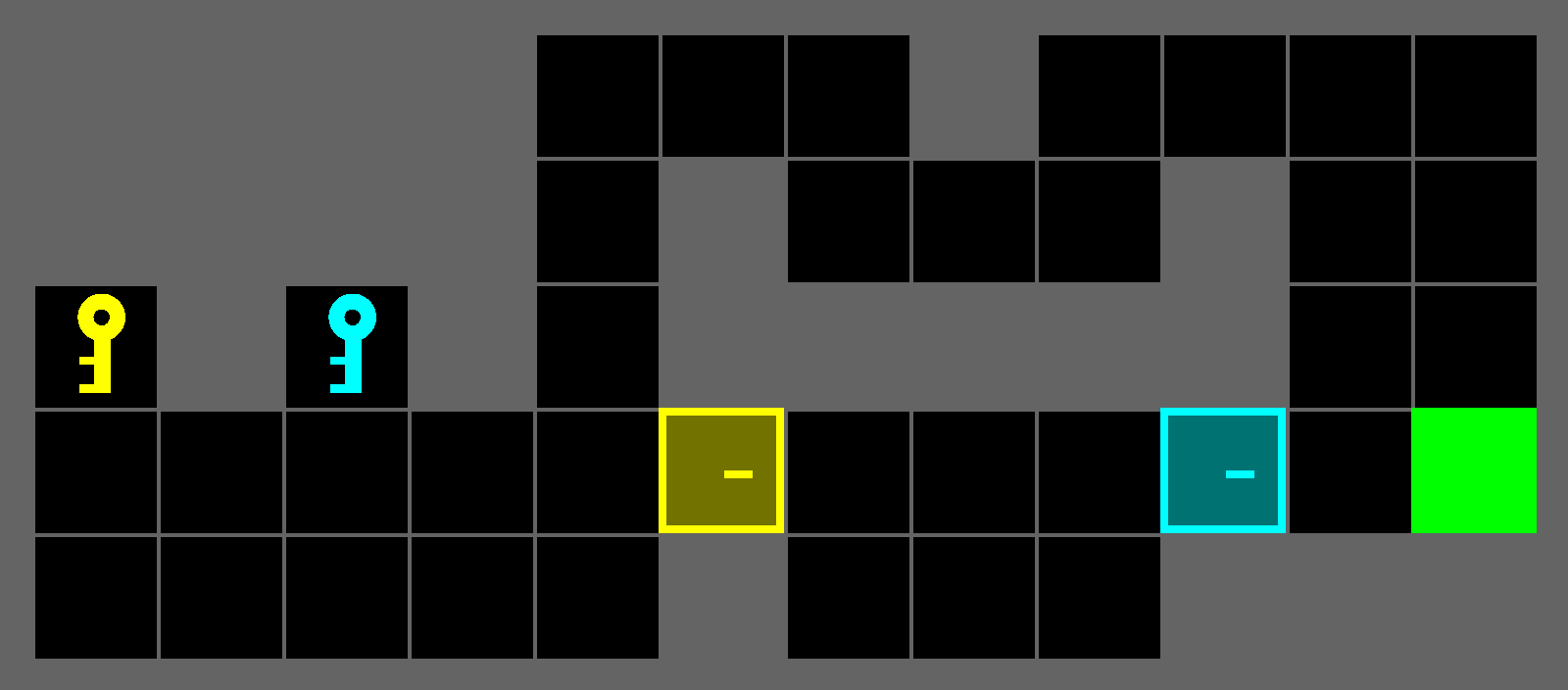}};

    \coordinate (ini) at (-110pt, -20pt);
    \coordinate (key) at (-110pt, 0pt);
    \coordinate (mni) at (-70pt, -20pt);
    \coordinate (mey) at (-70pt, 0pt);
    \coordinate (end) at (110pt, -20pt);

    \draw[
        {Circle[scale=0.75]}-{Triangle[scale=0.75]},
        red,
        line width=3pt,
        rounded corners=3pt]
        
        ($(ini) + (0, -6pt)$)
        -- ($(key) + (0, -7pt)$)
        -- ($(key) + (6pt, -7pt)$)
        -- ($(ini) + (6pt, 3pt)$)
        -- ($(mni) + (0, 3pt)$)
        -- ($(mey) + (0, -7pt)$)
        -- ($(mey) + (6pt, -7pt)$)
        -- ($(mni) + (6pt, 3pt)$)
        -- ($(end) + (-5pt, 3pt)$);

    \node[align=center] at (0, -69pt) {\large Optimal Trajectory};
\end{tikzpicture}

%% file: figures/tikz/gridworld-linked-constraint.tex
\begin{tikzpicture}
    \node at (0,0) {\includegraphics[width=250pt]{figures/raw/gridworld-linked}};

    \coordinate (ini) at (-110pt, -20pt);
    \coordinate (key) at (-110pt, 0pt);
    \coordinate (mni) at (-70pt, -20pt);
    \coordinate (mey) at (-70pt, 0pt);
    \coordinate (end) at (110pt, -20pt);

    \draw[
        {Circle[scale=0.75]}-{Triangle[scale=0.75]},
        red,
        line width=3pt,
        rounded corners=3pt]
        
        ($(ini) + (-4pt, 0)$)

            -- ($(-30pt, -20pt) + (0, 0)$)
            -- ($(-30pt, 40pt) + (0, 0)$)
            -- ($(10pt, 40pt) + (0, 0)$)
            -- ($(10pt, 20pt) + (0, 0)$)
            -- ($(50pt, 20pt) + (0, 0)$)
            -- ($(50pt, 40pt) + (0, 0)$)
            -- ($(90pt, 40pt) + (0, 0)$)
            -- ($(90pt, -20pt) + (0, 0)$)
            
            -- ($(end) + (-5pt,0)$);

    \node[
        cross out,
        draw,
        red,
        line width=2pt] at ($(mey) + (0pt, -10pt)$) {};

    \node[align=center] at (0, -69pt) {\large Constraint on Picking up the Second Key};
\end{tikzpicture}

%% file: figures/tikz/gridworld-linked-scores.tex
\begin{tikzpicture}
    \begin{axis}[
        colormap/RdPu,
        colorbar,
        colorbar style={ymin=0},
        point meta min=-0.2,
        point meta max=1,
        axis equal image,
        xlabel={Set-up Decision},
        ylabel={Pay-off Decision},
        ylabel style={yshift=-3pt},
        xmin=-0.5, xmax=14.5,
        ymin=-0.5, ymax=14.5,
        xtick={0,1,2,3,4,5,6,7,8,9,10,11,12,13,14},
        xticklabels={
            \shortstack{1\\\textbf{~K\textsubscript{1}}},
            \shortstack{2\\\textbf{~K\textsubscript{1}}},
            3,
            4,
            \shortstack{5\\\textbf{~K\textsubscript{2}}},
            \shortstack{6\\\textbf{~K\textsubscript{2}}},
            7,
            8,
            \shortstack{9\\\textbf{~D\textsubscript{1}}},
            \shortstack{10\\\textbf{~~D\textsubscript{1}}},
            11,
            12,
            \shortstack{13\\\textbf{~~D\textsubscript{2}}},
            \shortstack{14\\\textbf{~~D\textsubscript{2}}},
            15},
        ytick={0,1,2,3,4,5,6,7,8,9,10,11,12,13,14},
        yticklabels={
            \textbf{(K\textsubscript{1})} 1\textsuperscript{st}\hspace*{2pt},
            \textbf{(K\textsubscript{1})} 2\textsuperscript{nd},
            3\textsuperscript{rd},
            4\textsuperscript{th},
            \textbf{(K\textsubscript{2})} 5\textsuperscript{th},
            \textbf{(K\textsubscript{2})} 6\textsuperscript{th},
            7\textsuperscript{th},
            8\textsuperscript{th},
            \textbf{(D\textsubscript{1})} ~~9\textsuperscript{th},
            \textbf{(D\textsubscript{1})} 10\textsuperscript{th},
            11\textsuperscript{th},
            12\textsuperscript{th},
            \textbf{(D\textsubscript{2})} 13\textsuperscript{th},
            \textbf{(D\textsubscript{2})} 14\textsuperscript{th},
            15\textsuperscript{th}},
        tick label style={font=\footnotesize},
        xmajorgrids=true,
        xtick pos=bottom,
        ytick pos=left,
        tick align=outside]
        
    \addplot[
        matrix plot*,
        point meta=explicit]
        file {figures/raw/gridworld-linked-scores.dat};
    \end{axis}
\end{tikzpicture}

%% file: figures/tikz/irl-gridworld.tex
\begin{tikzpicture}
    \begin{axis}[
        width=200pt, height=140pt, scale only axis,
        x dir=reverse,
        axis y line*=right,
        xlabel={Stochasticity (Temperature Parameter, $\beta^{-1}$)},
        ylabel={Error in Reward Function},
        yticklabel style={color=cirl2},
        ylabel style={color=cirl2},
        ytick style={color=cirl2},
        xtick={0.2, 0.4, 0.6, 0.8, 1.0, 1.2, 1.4, 1.6, 1.8},
        xticklabels={$\frac{1}{0.2}$, $\frac{1}{0.4}$, $\frac{1}{0.6}$, $\frac{1}{0.8}$, $\frac{1}{1.0}$, $\frac{1}{1.2}$, $\frac{1}{1.4}$, $\frac{1}{1.6}$, $\frac{1}{1.8}$},
        ytick={0.57, 0.59, 0.61, 0.63, 0.65},
        ymin=0.57, ymax=0.65,
        yticklabel style={
            /pgf/number format/fixed,
            /pgf/number format/precision=2,
            /pgf/number format/zerofill},
        grid=both]

        \addplot[draw=none,name path=lower] file {figures/raw/irl-gridworld-reward-lower.dat};
        \addplot[draw=none,name path=upper] file {figures/raw/irl-gridworld-reward-upper.dat};
        \addplot[fill=cirl2,opacity=0.1] fill between[of=lower and upper];

        \addplot[
            cirl2,
            ultra thick,
            mark=*,
            mark size=2.5pt,
            mark options={draw=white, line width=1pt}]
            file {figures/raw/irl-gridworld-reward-mean.dat};
        
    \end{axis}
    \begin{axis}[
        width=200pt, height=140pt, scale only axis,
        x dir=reverse,
        axis x line=none,
        axis y line*=left,
        ylabel={Error in Strategic Link Scores},
        yticklabel style={color=cirl1},
        ylabel style={color=cirl1},
        ytick style={color=cirl1},
        ytick={0, 0.0003, 0.0006, 0.0009, 0.0012},
        ymin=0, ymax=0.0012,
        scaled y ticks=false,
        yticklabel style={
            /pgf/number format/fixed,
            /pgf/number format/precision=4}]

        \addplot[draw=none,name path=lower] file {figures/raw/irl-gridworld-scores-lower.dat};
        \addplot[draw=none,name path=upper] file {figures/raw/irl-gridworld-scores-upper.dat};
        \addplot[fill=cirl1,opacity=0.1] fill between[of=lower and upper];

        \addplot[
            cirl1,
            ultra thick,
            mark=*,
            mark size=2.5pt,
            mark options={draw=white, line width=1pt}]
            file {figures/raw/irl-gridworld-scores-mean.dat};
        
    \end{axis}
\end{tikzpicture}

%% file: figures/tikz/irl-shortcuts.tex
\begin{tikzpicture}
    \begin{axis}[
        width=200pt, height=140pt, scale only axis,
        x dir=reverse,
        axis y line*=right,
        xlabel={Stochasticity (Temperature Parameter, $\beta^{-1}$)},
        ylabel={Error in Reward Function},
        yticklabel style={color=cirl2},
        ylabel style={color=cirl2},
        ytick style={color=cirl2},
        xtick={1., 2., 3., 4., 5., 6., 7., 8., 9.},
        xticklabels={$\frac{1}{1.0}$, $\frac{1}{2.0}$, $\frac{1}{3.0}$, $\frac{1}{4.0}$, $\frac{1}{5.0}$, $\frac{1}{6.0}$, $\frac{1}{7.0}$, $\frac{1}{8.0}$, $\frac{1}{9.0}$},
        ytick={0.42, 0.48, 0.54, 0.60, 0.66},
        ymin=0.42, ymax=0.66,
        yticklabel style={
            /pgf/number format/fixed,
            /pgf/number format/precision=2,
            /pgf/number format/zerofill},
        grid=both]

        \addplot[draw=none,name path=lower] file {figures/raw/irl-shortcuts-reward-lower.dat};
        \addplot[draw=none,name path=upper] file {figures/raw/irl-shortcuts-reward-upper.dat};
        \addplot[fill=cirl2,opacity=0.1] fill between[of=lower and upper];

        \addplot[
            cirl2,
            ultra thick,
            mark=*,
            mark size=2.5pt,
            mark options={draw=white, line width=1pt}]
            file {figures/raw/irl-shortcuts-reward-mean.dat};
        
    \end{axis}
    \begin{axis}[
        width=200pt, height=140pt, scale only axis,
        x dir=reverse,
        axis x line=none,
        axis y line*=left,
        ylabel={Error in Strategic Link Scores},
        yticklabel style={color=cirl1},
        ylabel style={color=cirl1},
        ytick style={color=cirl1},
        ytick={0, 0.0004, 0.0008, 0.0012, 0.0016},
        ymin=0, ymax=0.0016,
        scaled y ticks=false,
        yticklabel style={
            /pgf/number format/fixed,
            /pgf/number format/precision=4}]

        \addplot[draw=none,name path=lower] file {figures/raw/irl-shortcuts-scores-lower.dat};
        \addplot[draw=none,name path=upper] file {figures/raw/irl-shortcuts-scores-upper.dat};
        \addplot[fill=cirl1,opacity=0.1] fill between[of=lower and upper];

        \addplot[
            cirl1,
            ultra thick,
            mark=*,
            mark size=2.5pt,
            mark options={draw=white, line width=1pt}]
            file {figures/raw/irl-shortcuts-scores-mean.dat};
        
    \end{axis}
\end{tikzpicture}

%% file: figures/tikz/irl-arthighway.tex
\begin{tikzpicture}
    \begin{axis}[
        width=200pt, height=140pt, scale only axis,
        x dir=reverse,
        axis y line*=right,
        xlabel={Stochasticity (Temperature Parameter, $\beta^{-1}$)},
        ylabel={Error in Reward Function},
        yticklabel style={color=cirl2},
        ylabel style={color=cirl2},
        ytick style={color=cirl2},
        xtick={.1, .2, .3, .4, .5, .6, .7, .8, .9},
        xticklabels={$\frac{1}{0.1}$, $\frac{1}{0.2}$, $\frac{1}{0.3}$, $\frac{1}{0.4}$, $\frac{1}{0.5}$, $\frac{1}{0.6}$, $\frac{1}{0.7}$, $\frac{1}{0.8}$, $\frac{1}{0.9}$},
        ytick={0.55, 0.60, 0.65, 0.70, 0.75},
        ymin=0.55, ymax=0.75,
        yticklabel style={
            /pgf/number format/fixed,
            /pgf/number format/precision=2,
            /pgf/number format/zerofill},
        grid=both]

        \addplot[draw=none,name path=lower] file {figures/raw/irl-arthighway-reward-lower.dat};
        \addplot[draw=none,name path=upper] file {figures/raw/irl-arthighway-reward-upper.dat};
        \addplot[fill=cirl2,opacity=0.1] fill between[of=lower and upper];

        \addplot[
            cirl2,
            ultra thick,
            mark=*,
            mark size=2.5pt,
            mark options={draw=white, line width=1pt}]
            file {figures/raw/irl-arthighway-reward-mean.dat};
        
    \end{axis}
    \begin{axis}[
        width=200pt, height=140pt, scale only axis,
        x dir=reverse,
        axis x line=none,
        axis y line*=left,
        ylabel={Error in Strategic Link Scores},
        yticklabel style={color=cirl1},
        ylabel style={color=cirl1},
        ytick style={color=cirl1},
        ytick={0, 0.01, 0.02, 0.03, 0.04},
        ymin=0, ymax=0.04,
        scaled y ticks=false,
        yticklabel style={
            /pgf/number format/fixed,
            /pgf/number format/precision=2}]

        \addplot[draw=none,name path=lower] file {figures/raw/irl-arthighway-scores-lower.dat};
        \addplot[draw=none,name path=upper] file {figures/raw/irl-arthighway-scores-upper.dat};
        \addplot[fill=cirl1,opacity=0.1] fill between[of=lower and upper];

        \addplot[
            cirl1,
            ultra thick,
            mark=*,
            mark size=2.5pt,
            mark options={draw=white, line width=1pt}]
            file {figures/raw/irl-arthighway-scores-mean.dat};
        
    \end{axis}
\end{tikzpicture}

%% file: figures/tikz/shortcuts.tex
\begin{tikzpicture}[
    ->,
    >=stealth,
    shorten >=1pt,
    node distance=1.60cm,
    semithick,
    draw=black!20,
    state/.append style={minimum size=0,inner sep=2pt}]

    \node[state] (node1) {1};
    \node[state] (node2) [right of=node1] {2};
    \node[state] (node3) [right of=node2] {3};
    \node[state] (node4) [right of=node3] {4};
    \node[state] (node5) [right of=node4] {5};

    \path[->] (node1) edge node[below,align=center] {\texttt{move}\\[-2pt]\scriptsize($-1$)} (node2);
    \path[->] (node2) edge node[below,align=center] {\texttt{move}\\[-2pt]\scriptsize($-1$)} (node3);
    \path[->] (node3) edge node[below,align=center] {\texttt{move}\\[-2pt]\scriptsize($-1$)} (node4);
    \path[->] (node4) edge node[below,align=center] {\texttt{move}\\[-2pt]\scriptsize($-1$)} (node5);

    \node at ($(node5)+(0,-15pt)$) {\scriptsize($+5$)};

    \path[->] (node1) 
        edge [in=105,out=165,looseness=8]
        node [above,yshift=1pt,align=center] {
            \textcolor{cpolimp1}{\texttt{prep}\textsubscript{1}}\\
            \textcolor{cpolimp3}{\texttt{prep}\textsubscript{2}}\\
            \textcolor{cpolimp2}{\texttt{prep}\textsubscript{3}}\\
            \textcolor{cpolimp4}{\texttt{prep}\textsubscript{4}}\\[-2pt]\scriptsize($-C$)} (node1);

    \path[->] (node2)
        edge[bend left=60,looseness=1.2]
        node[above,align=center] {
            \texttt{jump} \textit{\small if} \textcolor{cpolimp2}{\texttt{prep}\textsubscript{3}}\\[-2pt]\scriptsize$-(n\!=\!3)+(k\!=\!1)\!\cdot\!2C$} (node5);

    \path[->] (node1) 
        edge[bend left=30,looseness=1.1]
        node[above,align=center] {
            \texttt{jump} \textit{\small if} \textcolor{cpolimp4}{\texttt{prep}\textsubscript{4}}\\[-2pt]\scriptsize$-(n\!=\!2)+(k\!=\!1)\!\cdot\!2C$} (node3);

    \path[->] (node3)
        edge[bend left=30,looseness=1.1]
        node[above,align=center] {
            \texttt{jump} \textit{\small if} \textcolor{cpolimp1}{\texttt{prep}\textsubscript{1}} {\small $\wedge$} \textcolor{cpolimp3}{\texttt{prep}\textsubscript{2}}\\[-2pt]\scriptsize$-(n\!=\!2)+(k\!=\!2)\!\cdot\!2C$} (node5);

\end{tikzpicture}

%% file: figures/tikz/policy-improvement.tex
\begin{tikzpicture}
    \begin{axis}[
        width=200pt, height=150pt, scale only axis,
        xlabel={\small Number of Recommendations Implemented},
        ylabel={\small Final Performance},
        xmin = -0.2, xmax = 5.2, ymax=1.75,
        yticklabel style={
            /pgf/number format/fixed,
            /pgf/number format/precision=1,
            /pgf/number format/zerofill},
        tick label style={font=\small\sffamily},
        grid=both,
        legend style={
            at={(0.02,0.98)},
            anchor=north west,
            nodes={anchor=west,font={\scriptsize\sffamily}},
            legend columns=3,
            inner sep=1.5pt, row sep=-1pt,
            /tikz/every even column/.append style={column sep=3pt}},
        legend image code/.code={
            \draw[mark repeat=3,mark phase=2]
            plot coordinates {(0pt,0pt) (7pt,0pt) (14pt,0pt)};}]

        \tikzset{
            avg line/.style={
                ultra thick,
                mark=*,
                mark size=2.5pt,
                mark options={draw=white, line width=1pt}},
            min line/.style={
                thick,
                dash pattern=on 2pt off 1pt,
                mark=triangle*,
                mark size=2.5pt,
                mark options={solid, rotate=180, draw=white, line width=0.5pt}}}

        \addlegendimage{empty legend} \addlegendentry{Pick and Choose:}
        \addlegendimage{cpolimp2,avg line} \addlegendentry{Avg.}
        \addlegendimage{cpolimp2,min line} \addlegendentry{Worst Case}
        \addlegendimage{empty legend} \addlegendentry{All or Nothing:}
        \addlegendimage{cpolimp3,avg line} \addlegendentry{Avg.}
        \addlegendimage{cpolimp3,min line} \addlegendentry{Worst Case}
        \addlegendimage{empty legend} \addlegendentry{\textbf{Strategy-Aware:}}
        \addlegendimage{cpolimp1,avg line} \addlegendentry{Avg.}
        \addlegendimage{cpolimp1,min line} \addlegendentry{Worst Case}

        \addplot[black] coordinates {(-0.2, 0.5653974755282559) (5.2, 0.5653974755282559)};

        \addplot[cpolimp2,min line] file {figures/raw/polimp-pick-and-choose-min.dat};
        \addplot[cpolimp2,avg line] file {figures/raw/polimp-pick-and-choose.dat};
        \addplot[cpolimp3,avg line] file {figures/raw/polimp-all-or-nothing.dat};
        \addplot[cpolimp1,min line] file {figures/raw/polimp-strategic-min.dat};
        \addplot[cpolimp1,avg line] file {figures/raw/polimp-strategic.dat};

    \end{axis}
\end{tikzpicture}

%% file: figures/tikz/arthighway.tex
\begin{tikzpicture}[
    roadh/.style={very thick,color=cpolimp3},
    roada/.style={very thick,color=cpolimp1},
    mid arrow/.style={
        postaction={
            decorate,
            decoration={
                markings,
                mark=at position 0.7 with {\arrow[scale=1.2]{Stealth}}}}},
    mid arrow further/.style={
        postaction={
            decorate,
            decoration={
                markings,
                mark=at position 0.8 with {\arrow[scale=1.2]{Stealth}}}}},
    point/.style={fill,circle,minimum size=4pt,inner sep=0pt},
    node distance=36pt]

    \node[point,label=below:J1] (J1) {};
    \node[point,label=below:J2,right of=J1] (J2) {};
    \node[point,label=below:J3,right of=J2] (J3) {};
    \node[right=9pt of J3] (JX) {\textcolor{cpolimp1}{$\bm{\cdots}$}};
    \node[point,label=below:J9,right=9pt of JX] (J9) {};
    \node[point,label=below:J10,right of=J9] (J10) {};

    \node[point,above of=J1] (H1) {};
    \node[point,above of=J2] (H2) {};
    \node[point,above of=J3] (H3) {};
    \node[right=9pt of H3] (HX) {\textcolor{cpolimp3}{$\bm{\cdots}$}};
    \node[point,above of=J9] (H9) {};
    \node[point,above of=J10] (H10) {};

    \draw[roada,mid arrow] (J1) -- (J2);
    \draw[roada,mid arrow] (J2) -- (J3);
    \draw[roada] (J3) -- (JX);
    \draw[roada] (JX) -- (J9);
    \draw[roada,mid arrow] (J9) -- (J10);

    \draw[roadh,mid arrow] (H1) -- (H2);
    \draw[roadh,mid arrow] (H2) -- (H3);
    \draw[roadh] (H3) -- (HX);
    \draw[roadh] (HX) -- (H9);
    \draw[roadh,mid arrow] (H9) -- (H10);

    \draw[roadh,mid arrow] (J1) -- (H1);
    \draw[roadh,mid arrow] (J2) -- (H2);
    \draw[roadh,mid arrow] (J3) -- (H3);
    \draw[roadh,mid arrow] (J9) -- (H9);
    \draw[roadh,mid arrow] (J10) -- (H10);

    \node[left=18pt of J1] (ENT) {entry};
    \node[right=24pt of J10] (EXT) {exit};

    \draw[roadh,mid arrow further] (ENT) -- (J1);
    \draw[roada,mid arrow further] (J10) -- (EXT);
    \draw[roadh,bend left,mid arrow] (H10) to (EXT);

    \node[yshift=-21pt] at ($(J1)!0.5!(J10)$) {\textbf{\textcolor{cpolimp1}{Arterial Road}}};
    \node[yshift=11pt] at ($(H1)!0.5!(H10)$) {\textbf{\textcolor{cpolimp3}{Highway}}};
\end{tikzpicture}

%% file: figures/tikz/cars-sim-vehicles.tex
\begin{tikzpicture}
    \begin{axis}[
        width=200pt, height=160pt, scale only axis,
        xlabel={\small Time},
        xlabel style={yshift=9pt},
        ylabel={\small Cumulative Number of Vehicles},
        ymin=-2500, ymax=70000,
        xtick={0, 10000, 20000, 30000, 40000, 50000},
        xticklabels={0, \shortstack{10k\\\scriptsize\textbf{(intervention)}}, 20k, 30k, 40k, 50k},
        ytick={0, 10000, 20000, 30000, 40000, 50000, 60000},
        yticklabels={0, 10k, 20k, 30k, 40k, 50k, 60k},
        scaled x ticks=false,
        scaled y ticks=false,
        tick label style={font=\small\sffamily},
        grid=both,
        legend style={
            at={(0.02,0.98)},
            anchor=north west,
            nodes={anchor=west,font={\scriptsize\sffamily}},
            inner sep=1.5pt, row sep=-1pt}]

        \tikzset{my line/.style={ultra thick}}
        
        \definecolor{c0}{HTML}{CF384D}
        \definecolor{c1}{HTML}{ED6345}
        \definecolor{c2}{HTML}{FA9A58}
        \definecolor{c3}{HTML}{FDCE7C}
        \definecolor{c4}{HTML}{FEF1A7}
        \definecolor{c5}{HTML}{F3FAAD}
        \definecolor{c6}{HTML}{D1EC9C}
        \definecolor{c7}{HTML}{96D5A4}
        \definecolor{c8}{HTML}{5BB6A9}
        \definecolor{c9}{HTML}{3682BA}

        \draw[black,very thick] (axis cs:10000,-10000) -- (axis cs:10000,70000);

        \addplot[c0,my line] file {figures/raw/cars-sim-vehicles-j0.dat};
        \addplot[c1,my line] file {figures/raw/cars-sim-vehicles-j1.dat};
        \addplot[c2,my line] file {figures/raw/cars-sim-vehicles-j2.dat};
        \addplot[c3,my line] file {figures/raw/cars-sim-vehicles-j3.dat};
        \addplot[c4,my line] file {figures/raw/cars-sim-vehicles-j4.dat};
        \addplot[c5,my line] file {figures/raw/cars-sim-vehicles-j5.dat};
        \addplot[c6,my line] file {figures/raw/cars-sim-vehicles-j6.dat};
        \addplot[c7,my line] file {figures/raw/cars-sim-vehicles-j7.dat};
        \addplot[c8,my line] file {figures/raw/cars-sim-vehicles-j8.dat};
        \addplot[c9,my line] file {figures/raw/cars-sim-vehicles-j9.dat};

        \addlegendentry{J1, Arterial}
        \addlegendentry{J2, Arterial}
        \addlegendentry{J3, Arterial}
        \addlegendentry{J4, Arterial}
        \addlegendentry{J5, Arterial}
        \addlegendentry{J6, Arterial}
        \addlegendentry{J7, Arterial}
        \addlegendentry{J8, Arterial}
        \addlegendentry{J9, Arterial}
        \addlegendentry{J10, Arterial}
        
    \end{axis}
\end{tikzpicture}

%% file: figures/tikz/cars-sim-flow.tex
\begin{tikzpicture}
    \input{figures/tikz/cars}
    \begin{axis}[
        name=top,
        policy axis,
        title={\textbf{Pre-Intervention}},
        ylabel={\shortstack{Traffic Flow\\\scriptsize (Vehicles per Time Step)}},
        legend style={at={(0.98,0.98)}, anchor=north east}]

        \addplot+[csim1] file {figures/raw/cars-sim-flow-pre-arterial.dat};
        \addplot+[csim2] file {figures/raw/cars-sim-flow-pre-highway.dat};
    
        \addlegendentry{Arterial}
        \addlegendentry{Highway}
        
    \end{axis}
    \begin{axis}[
        at={(top.south)}, anchor=north, yshift=-30pt,
        policy axis,
        title={\textbf{Post-Intervention}},
        ylabel={\shortstack{Traffic Flow\\\scriptsize (Vehicles per Time Step)}},
        legend style={at={(0.98,0.98)}, anchor=north east}]

        \addplot+[csim1] file {figures/raw/cars-sim-flow-post-arterial.dat};
        \addplot+[csim2] file {figures/raw/cars-sim-flow-post-highway.dat};
    
        \addlegendentry{Arterial}
        \addlegendentry{Highway}
            
    \end{axis}
\end{tikzpicture}

%% file: figures/tikz/cars-sim-policy.tex
\begin{tikzpicture}
    \input{figures/tikz/cars}
    \begin{axis}[
        name=top,
        policy axis,
        title={\textbf{Pre-Intervention}},
        ylabel={\shortstack{Routing Policy\\\scriptsize (Freq. of Road given JX)}}]

        \addplot+[csim1] file {figures/raw/cars-sim-policy-pre-arterial.dat};
        \addplot+[csim2] file {figures/raw/cars-sim-policy-pre-highway.dat};
    
        \addlegendentry{Arterial}
        \addlegendentry{Highway}
        
    \end{axis}
    \begin{axis}[
        at={(top.south)}, anchor=north, yshift=-30pt,
        policy axis,
        title={\textbf{Post-Intervention}},
        ylabel={\shortstack{Routing Policy\\\scriptsize (Freq. of Road given JX)}}]

        \addplot+[csim1] file {figures/raw/cars-sim-policy-post-arterial.dat};
        \addplot+[csim2] file {figures/raw/cars-sim-policy-post-highway.dat};
    
        \addlegendentry{Arterial}
        \addlegendentry{Highway}
        
    \end{axis}
\end{tikzpicture}

%% file: figures/tikz/cars-sim-strategy.tex
\begin{tikzpicture}
    \input{figures/tikz/cars}
    \begin{axis}[
        strategy axis,
        ylabel={Strategic Link Score},
        ylabel style={yshift=-9pt}]
        
        \addplot+[csim3] file {figures/raw/cars-sim-strategy.dat};
    \end{axis}
\end{tikzpicture}

%% file: figures/tikz/cars-rl-policy.tex
\begin{tikzpicture}
    \input{figures/tikz/cars}
    \begin{axis}[
        name=top,
        policy axis,
        title={\textbf{Pre-Intervention}},
        ylabel={\shortstack{Routing Policy\\\scriptsize (Freq. of Road given JX)}}]

        \addplot+[csim1] file {figures/raw/cars-rl-policy-pre-arterial.dat};
        \addplot+[csim2] file {figures/raw/cars-rl-policy-pre-highway.dat};
    
        \addlegendentry{Arterial}
        \addlegendentry{Highway}
        
    \end{axis}
    \begin{axis}[
        at={(top.south)}, anchor=north, yshift=-30pt,
        policy axis,
        title={\textbf{Pre-Intervention}},
        ylabel={\shortstack{Routing Policy\\\scriptsize (Freq. of Road given JX)}}]

        \addplot+[csim1] file {figures/raw/cars-rl-policy-post-arterial.dat};
        \addplot+[csim2] file {figures/raw/cars-rl-policy-post-highway.dat};
    
        \addlegendentry{Arterial}
        \addlegendentry{Highway}
            
    \end{axis}
\end{tikzpicture}

%% file: figures/tikz/cars-rl-strategy.tex
\begin{tikzpicture}
    \input{figures/tikz/cars}
    \begin{axis}[
        strategy axis,
        ylabel={Strategic Link Score}]

        \addplot+[csim3] file {figures/raw/cars-rl-strategy.dat};
    \end{axis}
\end{tikzpicture}

%% file: main.bbl
\begin{thebibliography}{30}
\providecommand{\natexlab}[1]{#1}
\providecommand{\url}[1]{#1}
\csname url@samestyle\endcsname
\providecommand{\newblock}{\relax}
\providecommand{\bibinfo}[2]{#2}
\providecommand{\BIBentrySTDinterwordspacing}{\spaceskip=0pt\relax}
\providecommand{\BIBentryALTinterwordstretchfactor}{4}
\providecommand{\BIBentryALTinterwordspacing}{\spaceskip=\fontdimen2\font plus
\BIBentryALTinterwordstretchfactor\fontdimen3\font minus \fontdimen4\font\relax}
\providecommand{\BIBforeignlanguage}[2]{{%
\expandafter\ifx\csname l@#1\endcsname\relax
\typeout{** WARNING: IEEEtranSN.bst: No hyphenation pattern has been}%
\typeout{** loaded for the language `#1'. Using the pattern for}%
\typeout{** the default language instead.}%
\else
\language=\csname l@#1\endcsname
\fi
#2}}
\providecommand{\BIBdecl}{\relax}
\BIBdecl

\bibitem[Abbeel and Ng(2004)]{abbeel2004apprenticeship}
Abbeel, P. and A.~Y. Ng, ``Apprenticeship learning via inverse reinforcement learning,'' in \emph{International Conference on Machine Learning}, 2004.

\bibitem[Amir and Amir(2018)]{amir2018highlights}
Amir, D. and O.~Amir, ``{HIGHLIGHTS}: Summarizing agent behavior to people,'' in \emph{International Conference on Autonomous Agents and Multiagent Systems}, 2018.

\bibitem[Chen et~al.(2022)Chen, Silvestri, Tolomei, Wang, Zhu, and Ahn]{chen2022explain}
Chen, Z., F.~Silvestri, G.~Tolomei, J.~Wang, H.~Zhu, and H.~Ahn, ``Explain the explainer: Interpreting model-agnostic counterfactual explanations of a deep reinforcement learning agent,'' \emph{IEEE Transactions on Artificial Intelligence}, vol.~5, no.~4, pp. 1443--1457, 2022.

\bibitem[Cruz et~al.(2023)Cruz, Dazeley, Vamplew, and Moreira]{cruz2023explainable}
Cruz, F., R.~Dazeley, P.~Vamplew, and I.~Moreira, ``Explainable robotic systems: Understanding goal-driven actions in a reinforcement learning scenario,'' \emph{Neural Computing and Applications}, vol.~35, no.~25, pp. 18\,113--18\,130, 2023.

\bibitem[Ehsan et~al.(2018)Ehsan, Harrison, Chan, and Riedl]{ehsan2018rationalization}
Ehsan, U., B.~Harrison, L.~Chan, and M.~O. Riedl, ``Rationalization: A neural machine translation approach to generating natural language explanations,'' in \emph{AAAI/ACM Conference on AI, Ethics, and Society}, 2018.

\bibitem[Erwig et~al.(2018)Erwig, Fern, Murali, and Koul]{erwig2018explaining}
Erwig, M., A.~Fern, M.~Murali, and A.~Koul, ``Explaining deep adaptive programs via reward decomposition,'' in \emph{IJCAI/ECAI Workshop on Explainable Artificial Intelligence}, 2018.

\bibitem[Gleave et~al.(2021)Gleave, Dennis, Legg, Russell, and Leike]{gleave2020quantifying}
Gleave, A., M.~Dennis, S.~Legg, S.~Russell, and J.~Leike, ``Quantifying differences in reward functions,'' in \emph{International Conference on Learning Representations}, 2021.

\bibitem[Greydanus et~al.(2018)Greydanus, Koul, Dodge, and Fern]{greydanus2018visualizing}
Greydanus, S., A.~Koul, J.~Dodge, and A.~Fern, ``Visualizing and understanding {Atari} agents,'' in \emph{International Conference on Machine Learning}, 2018.

\bibitem[Haarnoja et~al.(2017)Haarnoja, Tang, Abbeel, and Levine]{haarnoja2017reinforcement}
Haarnoja, T., H.~Tang, P.~Abbeel, and S.~Levine, ``Reinforcement learning with deep energy-based policies,'' in \emph{International Conference on Machine Learning}, 2017.

\bibitem[Hein et~al.(2018)Hein, Udluft, and Runkler]{hein2018interpretable}
Hein, D., S.~Udluft, and T.~A. Runkler, ``Interpretable policies for reinforcement learning by genetic programming,'' \emph{Engineering Applications of Artificial Intelligence}, vol.~76, pp. 158--169, 2018.

\bibitem[Huber et~al.(2023)Huber, Demmler, Mertes, Olson, and Andr{\'e}]{huber2023ganterfactual}
Huber, T., M.~Demmler, S.~Mertes, M.~L. Olson, and E.~Andr{\'e}, ``Ganterfactual-rl: Understanding reinforcement learning agents' strategies through visual counterfactual explanations,'' \emph{arXiv preprint arXiv:2302.12689}, 2023.

\bibitem[Iyer et~al.(2018)Iyer, Li, Li, Lewis, Sundar, and Sycara]{iyer2018transparency}
Iyer, R., Y.~Li, H.~Li, M.~Lewis, R.~Sundar, and K.~Sycara, ``Transparency and explanation in deep reinforcement learning neural networks,'' in \emph{AAAI/ACM Conference on AI, Ethics, and Society}, 2018.

\bibitem[Juozapaitis et~al.(2019)Juozapaitis, Koul, Fern, Erwig, and Doshi-Velez]{juozapaitis2019explainable}
Juozapaitis, Z., A.~Koul, A.~Fern, M.~Erwig, and F.~Doshi-Velez, ``Explainable reinforcement learning via reward decomposition,'' in \emph{IJCAI/ECAI Workshop on Explainable Artificial Intelligence}, 2019.

\bibitem[Khan et~al.(2009)Khan, Poupart, and Black]{khan2009minimal}
Khan, O., P.~Poupart, and J.~Black, ``Minimal sufficient explanations for factored markov decision processes,'' in \emph{International Conference on Automated Planning and Scheduling}, 2009.

\bibitem[Laroche et~al.(2019)Laroche, Trichelair, and Des~Combes]{laroche2019safe}
Laroche, R., P.~Trichelair, and R.~T. Des~Combes, ``Safe policy improvement with baseline bootstrapping,'' in \emph{International Conference on Machine Learning}, 2019.

\bibitem[Liu et~al.(2018)Liu, Schulte, Zhu, and Li]{liu2018toward}
Liu, G., O.~Schulte, W.~Zhu, and Q.~Li, ``Toward interpretable deep reinforcement learning with linear model {U}-trees,'' in \emph{Joint European Conference on Machine Learning and Knowledge Discovery in Databases}, 2018.

\bibitem[Madumal et~al.(2020)Madumal, Miller, Sonenberg, and Vetere]{madumal2020explainable}
Madumal, P., T.~Miller, L.~Sonenberg, and F.~Vetere, ``Explainable reinforcement learning through a causal lens,'' in \emph{Proceedings of the AAAI conference on artificial intelligence}, vol.~34, no.~03, 2020, pp. 2493--2500.

\bibitem[Olson et~al.(2021)Olson, Khanna, Neal, Li, and Wong]{olson2021counterfactual}
Olson, M.~L., R.~Khanna, L.~Neal, F.~Li, and W.-K. Wong, ``Counterfactual state explanations for reinforcement learning agents via generative deep learning,'' \emph{Artificial Intelligence}, vol. 295, p. 103455, 2021.

\bibitem[Seo(2025)]{seo2025joss}
Seo, T., ``{UXsim}: lightweight mesoscopic traffic flow simulator in pure {Python},'' \emph{Journal of Open Source Software}, 2025.

\bibitem[Sharma et~al.(2024)Sharma, Benac, Parbhoo, and Doshi-Velez]{sharma2024decision}
Sharma, A., L.~Benac, S.~Parbhoo, and F.~Doshi-Velez, ``Decision-point guided safe policy improvement,'' \emph{arXiv preprint arXiv:2410.09361}, 2024.

\bibitem[Shu et~al.(2018)Shu, Xiong, and Socher]{shu2018hierarchical}
Shu, T., C.~Xiong, and R.~Socher, ``Hierarchical and interpretable skill acquisition in multi-task reinforcement learning,'' in \emph{International Conference on Learning Representations}, 2018.

\bibitem[Silva et~al.(2020)Silva, Gombolay, Killian, Jimenez, and Son]{silva2020optimization}
Silva, A., M.~Gombolay, T.~Killian, I.~Jimenez, and S.-H. Son, ``Optimization methods for interpretable differentiable decision trees applied to reinforcement learning,'' in \emph{International Conference on Artificial Intelligence and Statistics}, 2020.

\bibitem[Sun et~al.(2023)Sun, H{\"u}y{\"u}k, Jarrett, and van~der Schaar]{sun2023accountability}
Sun, H., A.~H{\"u}y{\"u}k, D.~Jarrett, and M.~van~der Schaar, ``Accountability in offline reinforcement learning: Explaining decisions with a corpus of examples,'' \emph{Conference on Neural Information Processing Systems}, 2023.

\bibitem[Topin and Veloso(2019)]{topin2019generation}
Topin, N. and M.~Veloso, ``Generation of policy-level explanations for reinforcement learning,'' in \emph{AAAI Conference on Artificial Intelligence}, 2019.

\bibitem[van~der Waa et~al.(2018)van~der Waa, van Diggelen, van~den Bosch, and Neerincx]{van2018contrastive}
van~der Waa, J., J.~van Diggelen, K.~van~den Bosch, and M.~Neerincx, ``Contrastive explanations for reinforcement learning in terms of expected consequences,'' \emph{arXiv preprint arXiv:1807.08706}, 2018.

\bibitem[Verma et~al.(2018)Verma, Murali, Singh, Kohli, and Chaudhuri]{verma2018programmatically}
Verma, A., V.~Murali, R.~Singh, P.~Kohli, and S.~Chaudhuri, ``Programmatically interpretable reinforcement learning,'' in \emph{International Conference on Machine Learning}, 2018.

\bibitem[Wu et~al.(2022)Wu, Wu, Qiu, Wang, and Long]{wu2022supported}
Wu, J., H.~Wu, Z.~Qiu, J.~Wang, and M.~Long, ``Supported policy optimization for offline reinforcement learning,'' \emph{Conference on Neural Information Processing Systems}, 2022.

\bibitem[Yao et~al.(2022)Yao, Parbhoo, Pan, and Doshi-Velez]{yao2022policy}
Yao, J., S.~Parbhoo, W.~Pan, and F.~Doshi-Velez, ``Policy optimization with sparse global contrastive explanations,'' \emph{arXiv preprint arXiv:2207.06269}, 2022.

\bibitem[Yau et~al.(2020)Yau, Russell, and Hadfield]{yau2020did}
Yau, H., C.~Russell, and S.~Hadfield, ``What did you think would happen? explaining agent behaviour through intended outcomes,'' in \emph{Conference on Neural Information Processing Systems}, 2020.

\bibitem[Ziebart et~al.(2008)Ziebart, Maas, Bagnell, Dey, et~al.]{ziebart2008maximum}
Ziebart, B.~D., A.~L. Maas, J.~A. Bagnell, A.~K. Dey \emph{et~al.}, ``Maximum entropy inverse reinforcement learning.'' in \emph{AAAI Conference on Artificial Intelligence}, 2008.

\end{thebibliography}
